\documentclass{article}

 \usepackage[preprint]{neurips_2026}

\usepackage[utf8]{inputenc} 
\usepackage[T1]{fontenc}    
\usepackage{url}            
\usepackage{booktabs}       
\usepackage{amsfonts}       
\usepackage{amsmath}        
\usepackage{nicefrac}       
\usepackage{microtype}      

\usepackage{array}
\usepackage{multirow}
\usepackage{multicol}
\usepackage{enumitem}
\usepackage{caption}
\usepackage{subcaption}
\usepackage{pifont}
\usepackage{graphicx}
\usepackage[dvipsnames]{xcolor}
\usepackage[
    colorlinks = true,
    linkcolor = brown,
    urlcolor = brown,
    citecolor = brown,
    anchorcolor = brown,
]{hyperref}

\usepackage{tablefootnote}
\DeclareSymbolFont{extraup}{U}{zavm}{m}{n}
\DeclareMathSymbol{\varspade}{\mathalpha}{extraup}{81}
\DeclareMathSymbol{\varclub}{\mathalpha}{extraup}{84}
\DeclareMathSymbol{\varheart}{\mathalpha}{extraup}{86}
\DeclareMathSymbol{\vardiamond}{\mathalpha}{extraup}{87}

\newcommand{\cmark}{\textcolor{JungleGreen}{\ding{51}}}%

\definecolor{top5000L}{rgb}{0.9411764705882353, 0.5019607843137255, 0.5019607843137255}
\definecolor{top1000L}{rgb}{1.0, 0.0, 0.0}
\definecolor{top200L}{rgb}{0.5450980392156862, 0.0, 0.0}

\definecolor{top5000M}{rgb}{0.6901960784313725, 0.7686274509803922, 0.8705882352941177}
\definecolor{top1000M}{rgb}{0.2549019607843137, 0.4117647058823529, 0.8823529411764706}
\definecolor{top200M}{rgb}{0.0, 0.0, 1.0}

\definecolor{top}{rgb}{0.5803921568627451, 0.0, 0.8274509803921568}
\definecolor{bottom}{rgb}{0.19607843137254902, 0.803921568627451, 0.19607843137254902}

\title{Metropolis-Scale Road Network Datasets \\ for Fine-Grained Urban Traffic Modeling}

\author{%
Fedor Velikonivtsev\thanks{Equal contribution} \\
HSE University, Yandex Research \\
\texttt{fvelikon@yandex-team.ru} \\
\And
Oleg Platonov$^\ast$ \\
HSE University, Yandex Research \\
\texttt{olegplatonov@yandex-team.ru} \\
\And
Ekaterina Alimaskina$^\ast$ \\
BRAIn Lab, Yandex Research \\
\texttt{alimaskina@yandex-team.ru} \\
\And
Gleb Bazhenov$^\ast$ \\
HSE University, Yandex Research \\
\texttt{gv-bazhenov@yandex-team.ru} \\
\And
Liudmila Prokhorenkova\thanks{Corresponding author} \\
Yandex Research \\
\texttt{ostroumova-la@yandex-team.ru} \\
}

\begin{document}

\maketitle

\begin{abstract}
Modeling traffic dynamics is a critical challenge for urban computing, with applications from real-time traffic management to infrastructure planning. However, progress in this area is fundamentally constrained by a lack of large-scale public datasets that capture the subtle properties of real city road networks. Existing benchmarks are often limited by their small scale, reliance on sparse highway traffic sensors, absence of true road connectivity information, and lack of information about road properties. To address this issue, we introduce datasets representing fine-grained road networks of two major cities, which are unique in their scale (up to 100{,}000 road segments), use of real road connectivity, presence of time series measurements for both traffic speed and volume at a 5-minute resolution, and inclusion of rich static road attributes. These datasets enable in-depth analysis of spatiotemporal traffic patterns and can serve as benchmarks for various ML applications. As a practical demonstration of the utility of our datasets and the challenges they present, we use them for the task of traffic forecasting. The size of the real-world road networks in our datasets reveals significant scalability issues in current traffic forecasting models. To address them, we propose a simple and efficient baseline that not only scales to large road graphs but also achieves forecasting performance competitive with other established spatiotemporal models. We hope that the proposed datasets will serve as a foundational resource for a broad range of research in traffic modeling, urban computing, and smart city development.
\end{abstract}

\section{Introduction}
\label{sec:introduction}

Urban traffic is a fundamental component of daily life in any modern city that directly impacts our routines, affecting everything from commute time and package delivery schedules to the response speed of emergency vehicles. Thus, being able to effectively analyze and model this data is crucial for making smarter decisions about infrastructure planning, developing real-time traffic management systems, and improving the efficiency of logistics \citep{zheng2014urban}.

Despite this importance, academic research is hindered by a fundamental gap between the complexity of real-world traffic systems and the limitations of publicly available data. First, many established datasets for traffic modeling, such as \texttt{METR-LA} and \texttt{PEMS-BAY}, are derived from sparse loop detector sensors on highways, and their graph structures are heuristically constructed rather than based on true road connectivity \citep{li2018diffusion, yu2018spatio}. While more recent public datasets have incorporated real road network topologies from OpenStreetMap, they often rely on proxy signals from specific vehicle types \citep{kaiser2025spatio} or low-frequency accident counts \citep{nippani2023graph}, failing to capture the complete, high-frequency dynamics of city-wide traffic. Further, many relevant large-scale systems for traffic modeling \citep{derrow2021eta} are often developed on proprietary datasets within the industry, creating powerful applications that remain inaccessible and difficult for the broader research community to build upon or verify.

To address these issues, we introduce \texttt{city-traffic-M} and \texttt{city-traffic-L}, two large-scale datasets that contain ground-truth road networks in two major cities with their real traffic dynamics. Sourced from dense GPS signals, these datasets provide high-frequency (5-minute) measurements of both traffic speed and volume for every road segment. The unique combination of a real road graph, rich static features, and fine-grained, comprehensive traffic measurements provides a valuable new public resource for urban traffic analysis. The release of these datasets opens previously unavailable directions for research. For the first time, researchers can rigorously study the intricate connection between network topology and traffic flow, create more realistic urban traffic simulations, and evaluate models on data that reflects the scale and complexity of real-world production systems.

As a demonstration of their utility and the challenges they present, we use our datasets for the task of traffic forecasting. We show that the scale of the proposed datasets, with up to 100,000 nodes, exposes significant scalability limitations of standard spatiotemporal architectures, which become computationally prohibitive. To address this issue, we propose a simple and efficient approach to graph-based traffic forecasting that not only scales to large road graphs but also achieves forecasting performance competitive with other established spatiotemporal models.

In summary, our main contributions are as follows:
\begin{itemize}[leftmargin=14pt,topsep=1pt]
\setlength\itemsep{1pt}
\item We release two metropolis-scale datasets for fine-grained urban traffic analysis, combining real road network topology with high-frequency traffic speed and volume measurements collected from dense GPS signals.
\item We provide the examples of new research opportunities these datasets enable, from studying the connection between traffic dynamics and structural properties of road network to creating more realistic benchmarks for traffic forecasting.
\item We use these datasets to show that many current spatiotemporal models face significant scalability challenges on real-world urban traffic data, and propose a simple and efficient design that helps overcome this issue and can be treated as a practical baseline in future studies.
\end{itemize}

\section{Existing public datasets for traffic analysis}
\label{sec:existing-datasets}

The advancement of data-driven urban traffic analysis critically depends on high-quality, large-scale public datasets. These benchmarks provide a common foundation for uncovering complex real-world patterns, evaluating predictive models, and ultimately advancing urban traffic systems. Recent works show a clear progression towards more complex and realistic data, but we also find a persistent gap between the needs faced in various modern applications and the scope of analysis permitted by existing data sources.

Some of the most popular datasets for traffic modeling are \texttt{METR-LA} and \texttt{PEMS-BAY}, which were introduced by \citet{li2018diffusion}. They are primarily used for traffic forecasting with graph neural networks (GNNs) as one of the main predictive tasks related to traffic modeling. In these datasets, nodes represent sensors located on roads that measure traffic speed, and edges are constructed based on location proximity (road travel distance between the sensors). \texttt{METR-LA} is based on the data from loop detectors on the highways of Los Angeles County \citep{jagadish2014big}, and \texttt{PEMS-BAY} is based on the data from California Department of Transportation Performance Measurement System \citep{chen2001freeway}. Some works also use other datasets collected from the same PeMS data source: they may include different subsets of sensors or measurements during different periods of time, but the general structure of these datasets is mostly the same \citep{yu2018spatio, guo2019attention, song2020spatial}. Most works on traffic forecasting with GNNs evaluate their models exclusively on \texttt{METR-LA}, \texttt{PEMS-BAY}, or other datasets obtained from the PeMS data.

An important limitation of these standard datasets is the induced graph structure. It relies on very sparse distribution of sensors, which are typically located on intercity highways, and thus their measurements fail to capture complex urban traffic within cities. The sparsity of sensors also prevents the usage of real road network structure since the measurements are only available for a very small fraction of road segments. Thus, the real road networks are not provided with any of the standard datasets, and many previous works~\citep{li2018diffusion,yu2018spatio,wu2019graph,liu2023largest} connect two sensors if the road network distance between them is below a certain heuristically chosen threshold. In a more recent work, \citet{lin2025iutf} provide traffic measurements at sparse sensor locations spread across 40 cities (resulting in several hundred sensors per city), which is valuable for cross-city analysis but significantly differs from the complete and dense segment-level coverage of individual cities. \cite{zhang2024bjtt} introduce traffic measurements for a subset of major city roads, which is likewise a meaningful source of data but does not cover the full road network of the city and thus may prevent an in-depth analysis of how network structure impacts traffic flow.

\begin{figure*}[t!]
    \centering
    \begin{subfigure}{0.20\linewidth}
    \centering
    \includegraphics[width=\linewidth]{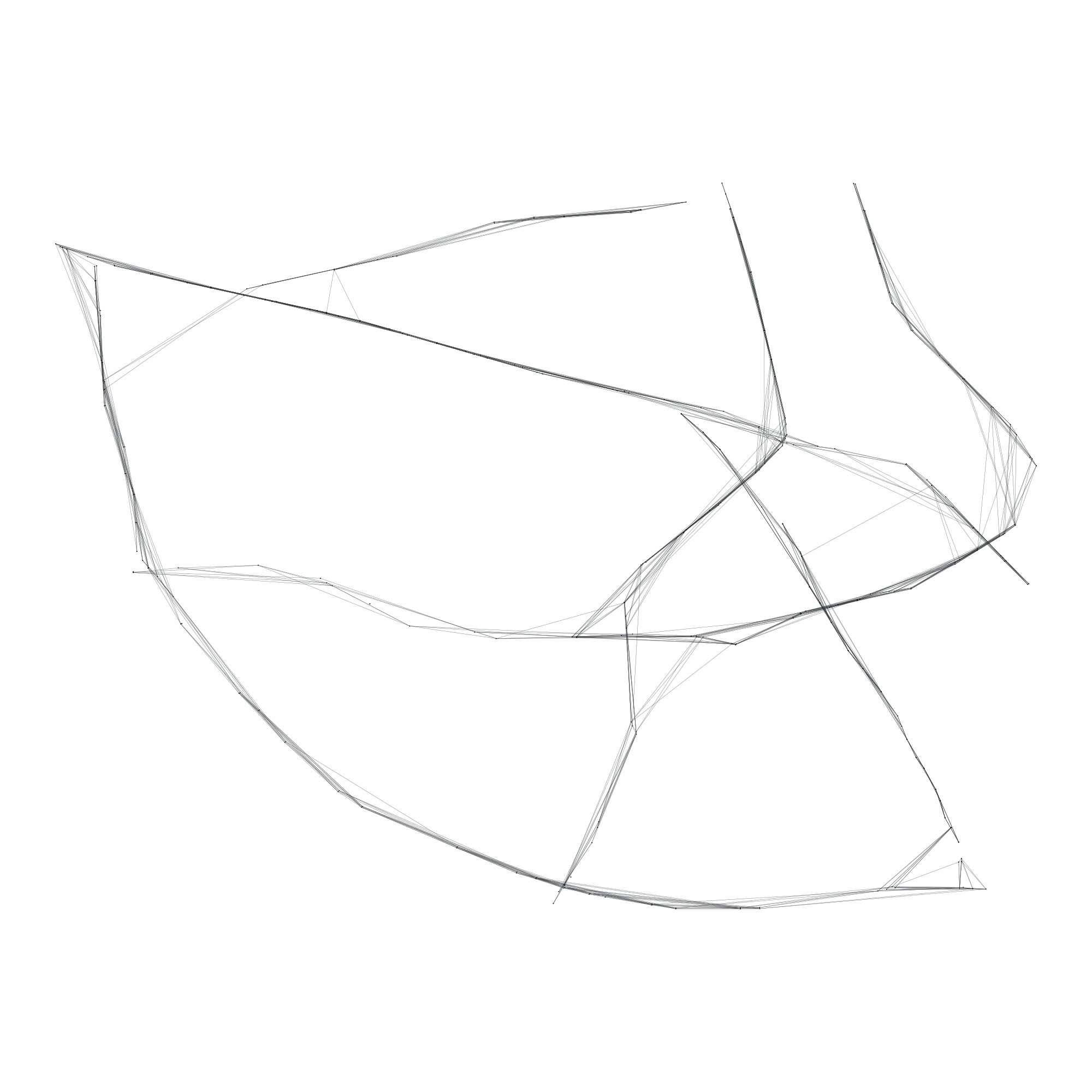}
    \caption{\texttt{PEMS-BAY}}
    \label{fig:pems-bay-layout}
    \end{subfigure}
    \hspace{20pt}
    \begin{subfigure}{0.24\linewidth}
    \centering
    \includegraphics[width=\linewidth]{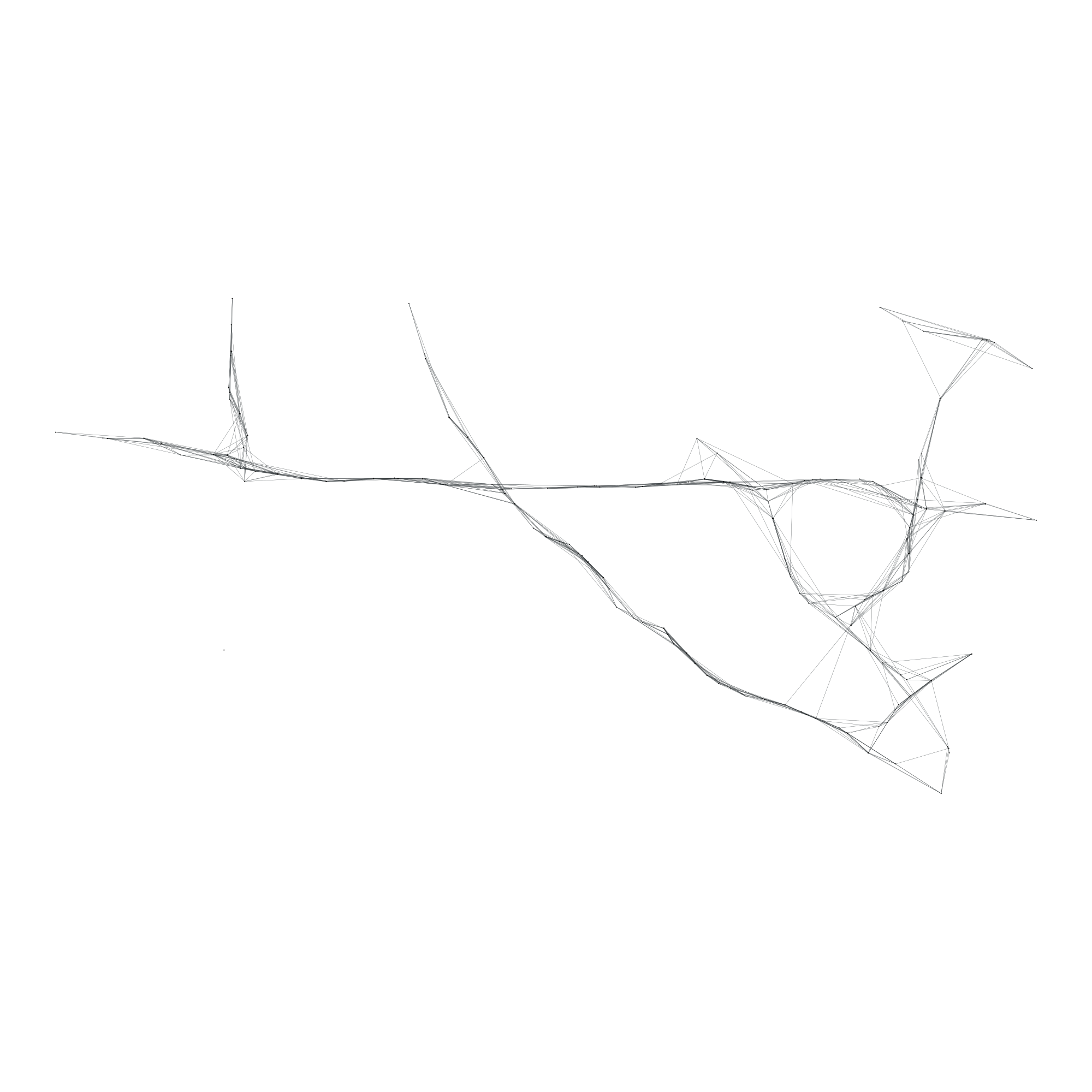}
    \caption{\texttt{METR-LA}}
    \label{fig:metr-la-layout}
    \end{subfigure}
    \begin{subfigure}{0.36\linewidth}
    \centering
    \includegraphics[width=\linewidth]{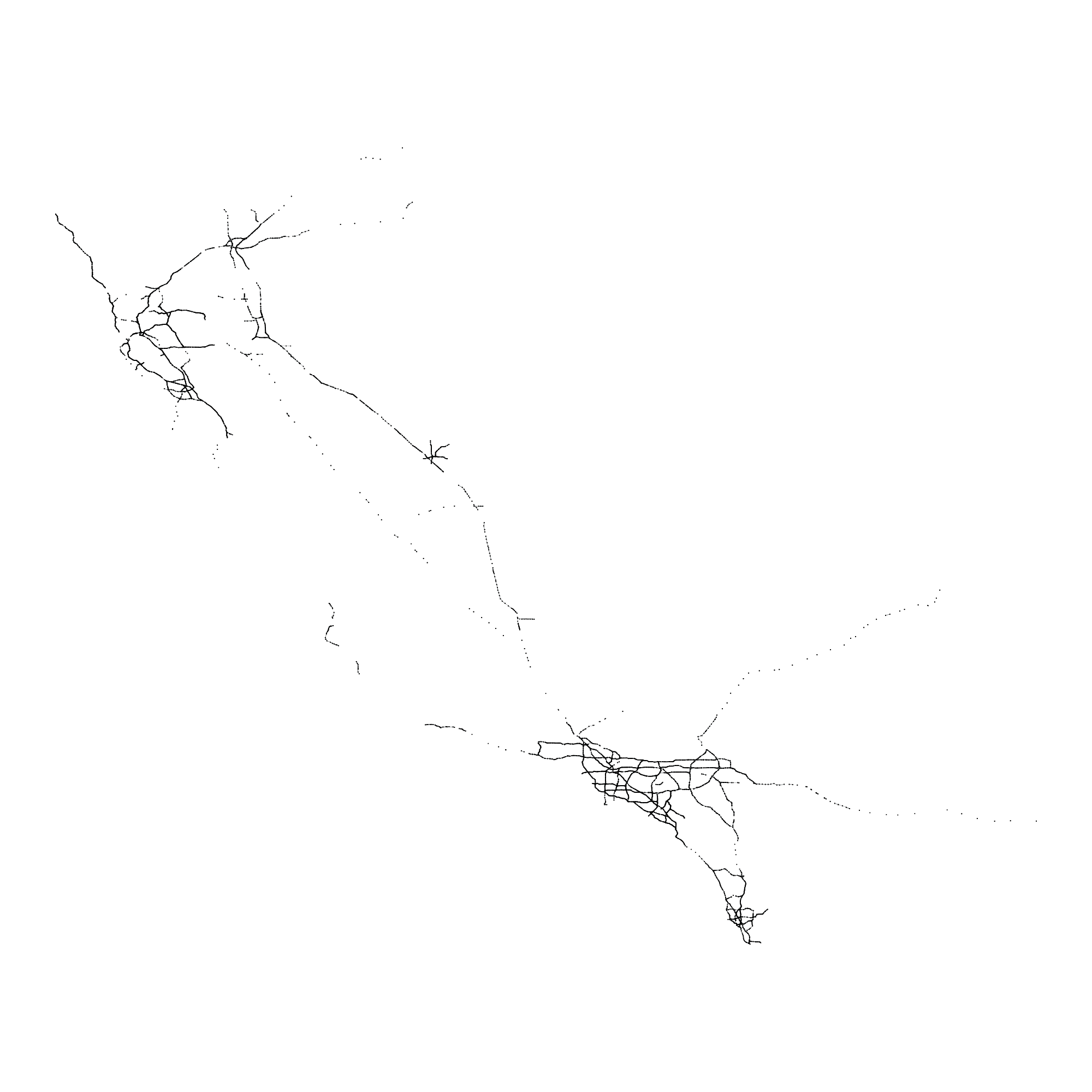}
    \caption{\texttt{LargeST}}
    \label{fig:large-st-layout}
    \end{subfigure}
    \begin{subfigure}{0.42\linewidth}
    \centering
    \includegraphics[width=\linewidth]{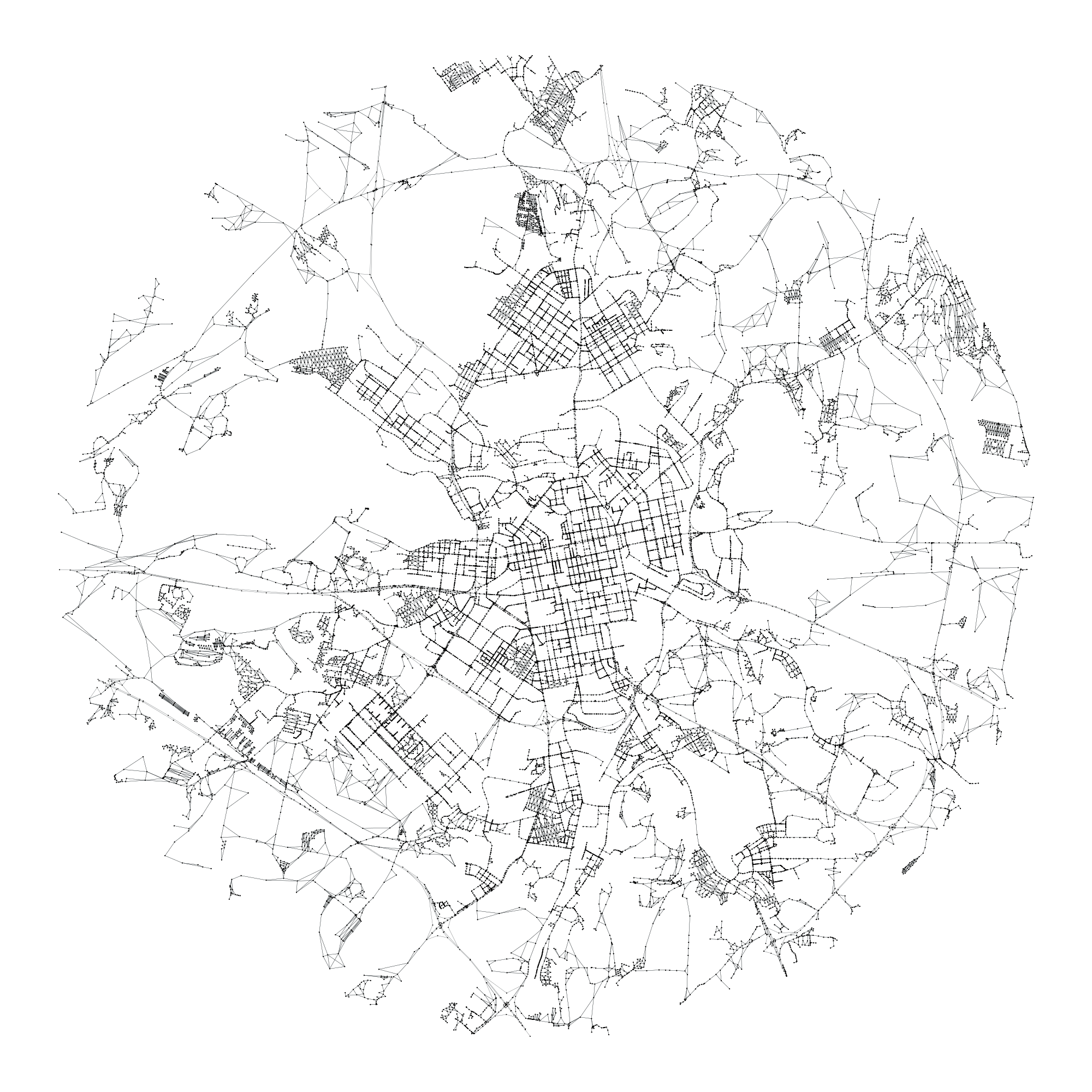}
    \caption{\texttt{city-traffic-M}}
    \label{fig:city-traffic-M-layout}
    \end{subfigure}
    \hspace{10pt}
    \begin{subfigure}{0.42\linewidth}
    \centering
    \includegraphics[width=\linewidth]{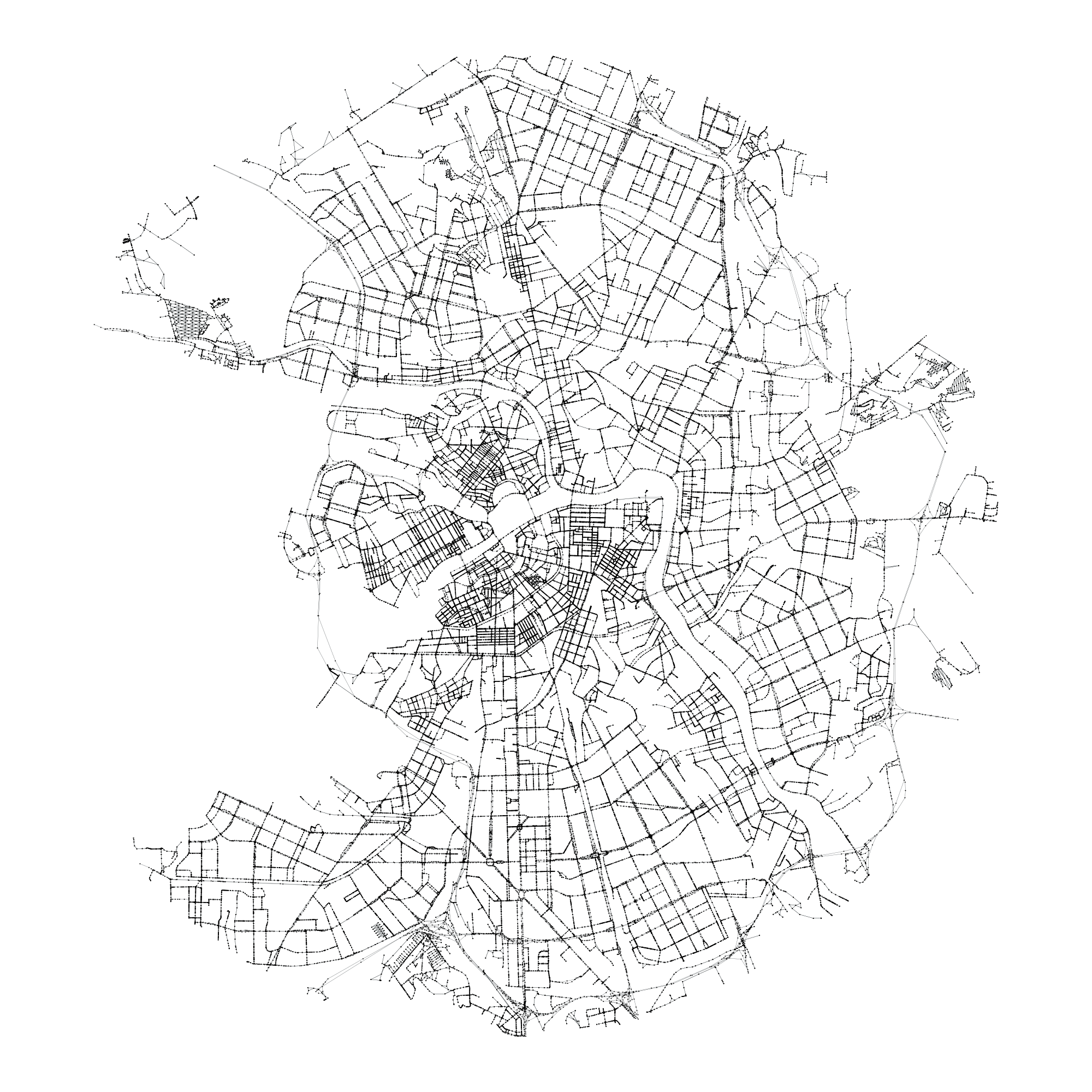}
    \caption{\texttt{city-traffic-L}}
    \label{fig:city-traffic-L-layout}
    \end{subfigure}
    \caption{Graph layout in some existing datasets (three leftmost) and the proposed \texttt{city-traffic} datasets (two rightmost). In \ref{fig:pems-bay-layout}, \ref{fig:metr-la-layout}, and \ref{fig:large-st-layout}, nodes correspond to sensors, graph structure is heuristically constructed based on road distances, layout is defined by sensor locations. In \ref{fig:city-traffic-M-layout} and \ref{fig:city-traffic-L-layout}, nodes correspond to distinct road segments, graph structure is defined by road adjacency according to traffic regulations, layout is defined by segment locations.}
    \label{fig:visualization-exampes}
    \vspace{-10pt}
\end{figure*}

There have been several attempts to address these issues by incorporating real road network topology, typically from OpenStreetMap (OSM). These datasets are designed for various urban traffic analysis tasks beyond standard forecasting. For instance, the datasets introduced by \citet{kaiser2025spatio} provide traffic volume on real road networks for the purpose of spatial interpolation. However, the estimates of traffic volume itself are derived from the taxi trip pickup and drop-off locations data, which does not allow for the precise representation of the entire traffic flow. Since having traffic data on every road segment is important but typically very challenging, \citet{xu2024unified} proposed a complex procedure to generate estimates of city-wide traffic based on sparse public data. Another work by \citet{nippani2023graph} offers a large-scale resource with 9 million records on real road networks from 8 US states for the purpose of accident prediction. While its scale and use of official reports are significant contributions, the primary temporal signal in this dataset is low-frequency accident counts, in contrast to more fine-grained measurements of speed and volume required for dynamic modeling. A distinct line of work leverages road networks not for modeling traffic phenomena, but for more fundamental graph machine learning research. For instance, \citet{liang2025towards} use large-scale OSM road networks to create a benchmark for quantifying long-range interactions in GNNs. Although it successfully utilizes the topology of real road networks, it remains unrelated to traffic modeling itself: the node labels are synthetically generated based on graph structural properties, and the dataset contains no real-world, temporal traffic measurements.

Thus, despite significant progress, no single publicly available resource combines all the necessary components for comprehensive, fine-grained urban traffic modeling. Existing datasets either rely on sparse highway sensors, derive traffic measurements from limited data available for specific vehicle types, generate estimates from incomplete data, provide only low-frequency signals like accident counts, or generate synthetic targets for tasks unrelated to traffic analysis. A critical gap remains for a benchmark that simultaneously provides metropolis-scale coverage of a dense urban road network, the ground-truth graph structure with rich road segment attributes, and high-frequency temporal measurements for both traffic speed and volume. Our work aims to address this issue, enabling researchers to create a more holistic approach to urban traffic modeling.

\section{New \texttt{city-traffic} datasets}
\label{sec:city-traffic-datasets}

\begin{figure*}[t!]
\centering
\includegraphics[width=\linewidth]{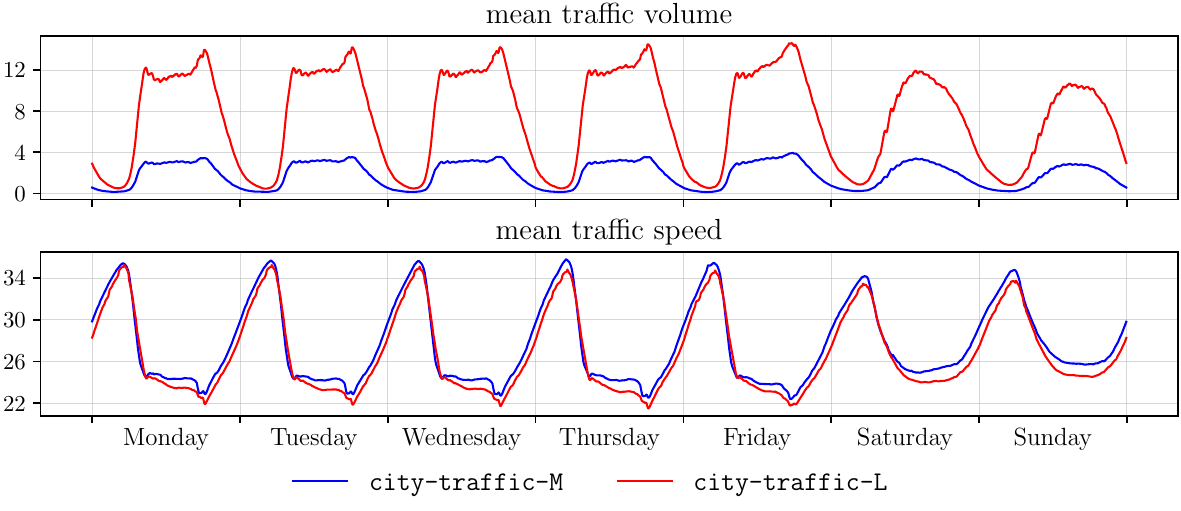}
\caption{Weekly dynamics of traffic variables volume and speed averaged across all road segments and weekly time periods in the proposed \texttt{city-traffic} datasets.}
\label{fig:weekly-dynamics}
\vspace{-10pt}
\end{figure*}

In our work, we present the first openly available datasets for large-scale and fine-grained study of urban traffic. We collect two spatiotemporal graph datasets from two major cities: \texttt{city-traffic-M} with more than 50{,}000 nodes and \texttt{city-traffic-L} with almost 100{,}000 nodes. These datasets differ significantly from the previous traffic modeling datasets in what the graphs represent and how they are constructed. While previous datasets only have information about real traffic registered at sparse sensor locations or provide some estimates generated using sparse public data, the information in our datasets was obtained from dense GPS signals transmitted by vehicles, and therefore the measurements are available at a fine-grained level of individual road segments. Thus, our graphs have nodes corresponding to \textit{all road segments in the two considered cities}. For each road segment, we provide two dynamic variables: current traffic speed and volume.\footnote{Traffic volume is derived from vehicles equipped with GPS, so its values underestimate the actual traffic flow, while preserving its dynamics. See Appendix~\ref{app:datasets-details} for more details about the proposed \texttt{city-traffic} datasets.} This data is provided at a 5-minute granularity, spanning from July 1st, 2024, to November 1st, 2024. For traffic speed, missing values can occur due to insufficient traffic volume for certain road segments at specific timestamps. For example, in \texttt{city-traffic-L}, the proportion of missing speed values can range from $5\%$ to $25\%$, depending on the time of day~--- a level of missingness inherent to real traffic data. For traffic volume, there are no missing values. In Figure~\ref{fig:weekly-dynamics}, we visualize the weekly traffic dynamics by averaging each variable across all road segments in a city. The resulting curves clearly exhibit daily traffic patterns: pronounced morning and evening rush hours on working days manifest an increase in traffic volume and a rapid decrease in average speed. On weekends, both variables evolve more gradually and typically show smaller variance. While the average speed profiles of \texttt{city-traffic-M} and \texttt{city-traffic-L} are similar, traffic volumes differ substantially, which reflects different levels of the overall demand between the two cities. Overall, the \texttt{city-traffic} datasets are the first to provide information on traffic volume and traffic speed simultaneously, which allows for a more holistic approach to traffic modeling. They represent a practical setting for traffic analysis by a real monitoring system, which contrasts with the previous datasets that only roughly approximate it due to incomplete data.

Further, while some widely-used previous datasets construct edges heuristically based on travel distance between sensors, graphs in \texttt{city-traffic} have edges representing actual road connectivity, which can provide much more information. In these graphs, a directed edge connects two road segments if they are incident to each other and moving from one segment to the other is permitted by traffic rules. Figure~\ref{fig:visualization-exampes} compares the graph structures of the datasets proposed in this work with the graph structure of some widely-used previous datasets. As can be seen, the graph structure in \texttt{city-traffic} provides a more complete and accurate representation of urban road networks.

Next, \texttt{city-traffic} datasets have rich node features describing the properties of road segments that can help investigate how road segment properties affect traffic dynamics. In particular, we provide 26 static attributes, including segment length, speed limit, coordinates of the segments' endpoints, quality of road surface, indicator of a mass transit lane, presence of a crosswalk, restriction for certain types of vehicles, etc. Technically, road attributes represent a mixture of numerical and categorical features. We show how some of the node features affect the traffic dynamics in Appendix~\ref{app:relation}.

\section{Data-driven analysis of urban traffic}
\label{sec:traffic-analysis}

As discussed above, most public traffic benchmarks have primarily focused on forecasting, providing time series at a limited number of sensor locations and relying on distance-based graph constructions commonly used in spatiotemporal traffic forecasting~\citep{li2018diffusion, yu2018spatio}. In contrast, our fine-grained metropolis-scale datasets open new opportunities for traffic modeling and analysis beyond forecasting. Many questions in transportation science and urban computing are related to the analysis of congestion, such as identifying traffic bottlenecks, estimating how delays concentrate across the network, analyzing travel time reliability~\citep{fhwa2017reliability, carrion2012reliability}, or investigating infrastructure criticality or network vulnerability \citep{jafino2020criticality}. While these directions are actively studied, analysis at a metropolis scale is often hard to reproduce due to limited public data or is otherwise unrealistic because of the traffic measurements being obtained through simulation~\citep{haghani2014datasource,jafino2020criticality}. Our new \texttt{city-traffic} datasets enable this line of work by providing the temporal measurements of both speed and volume at a 5-minute resolution for all road segments within the considered cities, along with their road network connectivity. Below, we provide an example of the analysis that can be conducted based on these datasets.

\subsection{Structural heterogeneity in urban road networks}

A common way to analyze large urban road networks is to identify the \emph{structural roles} of distinct road segments. The structural importance of roads can arise for different reasons: some segments are located in densely connected parts of the network, while others act as important bridges between otherwise weakly connected areas \citep{barthelemy2011spatial}. To capture these complementary roles, we use two standard centrality measures: PageRank \citep{page1999pagerank}, which quantifies how embedded a segment is in the network, and betweenness centrality \citep{crucitti2006centrality,kirkley2018betweenness}, which highlights segments lying on many shortest paths. Figure \ref{fig:centrality-visualization} in Appendix \ref{app:centrality-layouts} shows the road segments with the greatest values of these centrality measures in the two proposed \texttt{city-traffic} datasets. As can be seen, segments with the highest betweenness values often correspond to high-capacity arterial roads and act as important transition elements between distinct parts of the city. By contrast, segments with the highest PageRank values are not tied to long cross-city routes and rather appear in dense street meshes; they can be interpreted as distribution elements that organize traffic within urban districts and connect local streets to higher-level roads. This highlights an important limitation of commonly used settings: \emph{sensor-based datasets can miss fine-grained structural heterogeneity}. They rely on heuristic graph constructions that prevents the study of many local effects in road network connectivity.

\begin{figure*}[t!]
\centering
\includegraphics[width=0.8\linewidth]{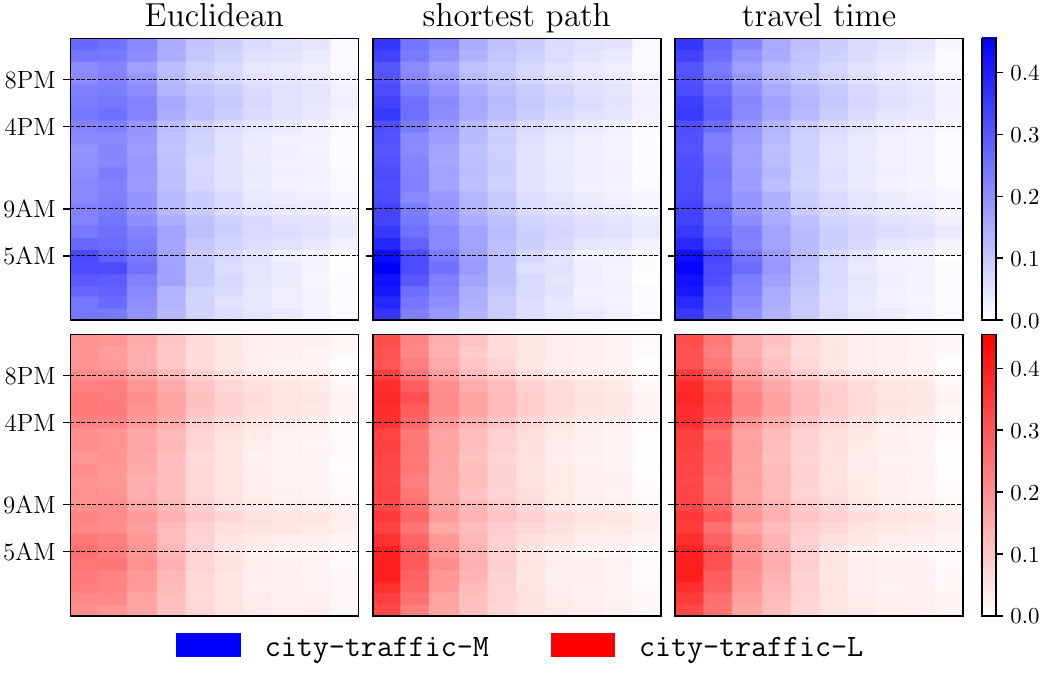}
\caption{Heatmaps with the mean correlation of segment-level delay $\Delta_s(t)$ between pairs of road segments aggregated by distance bin ($x$-axis) and local hour ($y$-axis) for three distance definitions: Euclidean, shortest path length, travel time. The $x$-axis orders segment pairs from shorter to longer distances. Dashed lines mark boundaries for representative rush-hour periods.}
\label{fig:delay-correlation}
\vspace{-10pt}
\end{figure*}

\subsection{Connecting congestion and reliability with road network structure}
\label{sec:congestion-and-structure}

Beyond forecasting, transportation engineering studies how congestion and reliability manifest across urban road networks in practice. Such analysis plays a central role in network monitoring and planning and typically relies on the notion of road segment \emph{problematicity}. It is rarely summarized by a single scalar quantity and instead characterized using a set of complementary traffic-based statistics that capture different operational aspects of traffic conditions, such as how frequently congestion occurs, how severe it becomes under adverse conditions, and how large its overall impact is when weighted by demand \citep{fhwa2017reliability,carrion2012reliability,seong2023congestion}. Applying such segment-level congestion and reliability measures at a metropolitan scale is challenging in practice, as it requires traffic observations consistently aligned with individual road segments on a realistic network. However, \emph{network topology and segment-level traffic measurements are rarely available jointly at a city scale}: traffic information is often only partially observed or provided via proprietary products or model-based pipelines, which hinders independent validation and limits systematic joint analysis of network structure and traffic dynamics \citep{haghani2014datasource,huang2023criticallinks}.

Our datasets enable such an analysis. Following the widely used practice in measuring reliability, we estimate the free-flow speed $v^{\text{free}}_s$ for each segment via the upper tail of the observed speed distribution and use the 95th percentile as a simple empirical proxy \citep{fhwa2017reliability}. Given this baseline, we characterize congestion using three representative segment-level indicators that are typically used in applied congestion and reliability analysis \citep{carrion2012reliability,seong2023congestion} to capture complementary effects of traffic conditions.

\begin{figure*}[t!]
\centering
\includegraphics[width=\linewidth]{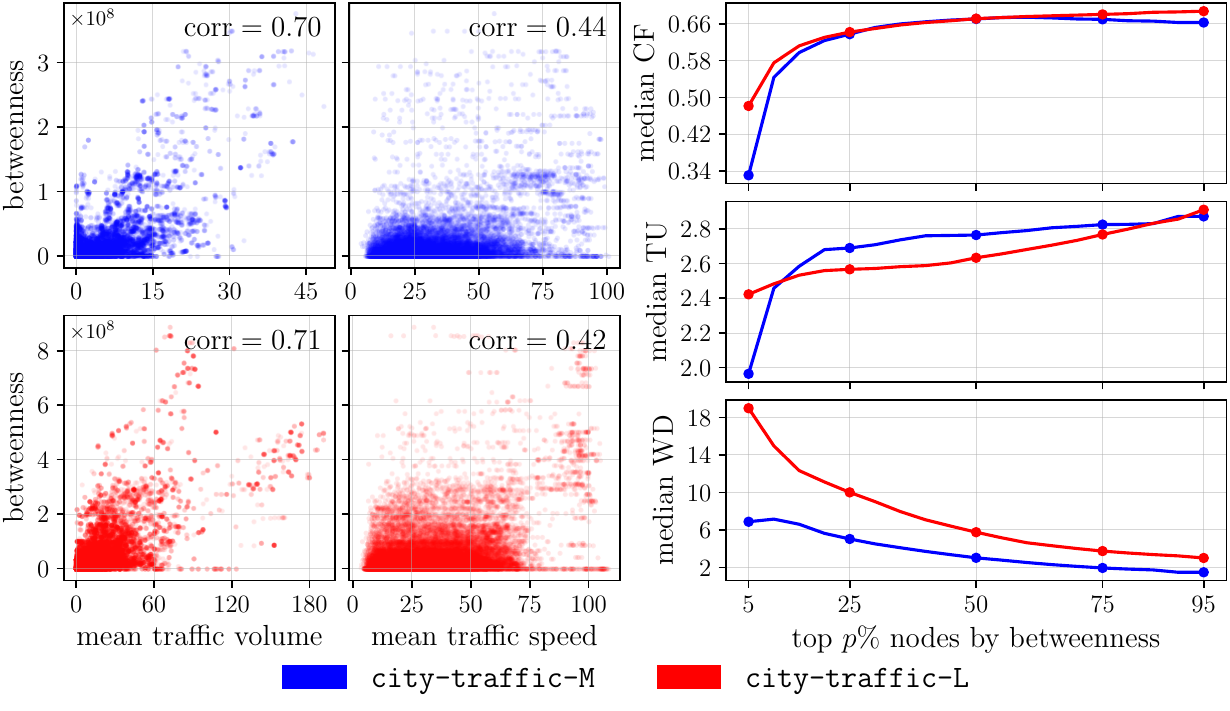}
\caption{Relation of node betweenness centrality with traffic variables and congestion measures in the proposed \texttt{city-traffic} datasets. \textbf{Left}: Scatter plots of betweenness centrality against mean traffic volume and speed (each panel is annotated with Pearson correlation). \textbf{Right}: Median value of congestion frequency (CF), tail unreliability (TU), and weighted delay (WD) among the top $p\%$ nodes ranked by betweenness.}
\label{fig:betweenness-vs-rest}
\vspace{-10pt}
\end{figure*}

\textbf{Congestion Frequency (CF)} measures how often a segment operates in congested conditions:
\[
\text{CF}(s) = \mathbb{E}_t\!\left[\mathbf{1}\{v_s(t) < \alpha v^{\text{free}}_s\}\right] \text{ for } \alpha = 0.8.
\]

\textbf{Tail Unreliability (TU)} captures the risk of rare but severe slowdowns via the travel-time index:
\[
\text{TU}(s) = 0.95\text{-quantile}_t\!\big[\mathrm{TTI}_s(t)\big], \text{ where } \mathrm{TTI}_s(t)=v^{\text{free}}_s/v_s(t).
\]

\textbf{Weighted Delay (WD)} weights excess travel time by observed volume to approximate cumulative congestion:
\[
\text{WD}(s) = \mathbb{E}_t\!\left[\mathrm{volume}_s(t)\cdot\Delta_s(t)\right], \text{ where } \Delta_s(t) = \mathrm{TTI}_s(t) - 1.
\]

First, we explore how traffic dependencies are aligned with the road network structure. We compute hourly Pearson correlations between segment-level delay series $\Delta_s(t)$ and aggregate them over distance bins defined by Euclidean distance, shortest path length, or free-flow network travel time. Figure~\ref{fig:delay-correlation} shows that correlations are generally higher for graph-aware distances than for Euclidean distance, especially among nearby segment pairs, which suggests that congestion dependencies follow ground-truth road connectivity more closely than straight-line spatial proximity. Another key observation is that during rush hours, delay correlation remains higher even for distant segment pairs, indicating that congestion effects propagate farther and persist longer in the road network.

We next analyze how the congestion indicators vary with structural importance. We exploit road betweenness centrality, a structural ranking heuristic that is widely used in applied criticality analysis \citep{jafino2020criticality}. Figure~\ref{fig:betweenness-vs-rest} (left) provides the scatter plots for this centrality measure against the two available traffic variables. As can be seen, betweenness centrality exhibits a strong positive correlation with both traffic volume and speed, which is natural as the roads with high betweenness are expected to carry substantial amounts of traffic volume. The same Figure \ref{fig:betweenness-vs-rest} (right) presents how the median value of congestion measures introduced above varies among the top $p\%$ nodes ranked by betweenness. The road segments selected by betweenness centrality exhibit significantly lower values of \emph{congestion frequency} and \emph{tail unreliability}, which is consistent with the fact that such segments often correspond to high-capacity arterial roads that typically operate with stable traffic volume and sustained high speed under normal conditions. It is also supported by Figure~\ref{fig:congestion-visualization}, which shows that the arterial roads are among the least problematic according to CF and TU measures. However, the values of the \emph{weighted delay} indicator, which reweighs the travel time index by traffic volume, are much higher for the candidate roads selected by betweenness. As discussed above, such roads are expected to carry substantial traffic volume, so the corresponding delays may significantly affect the WD measure. Further analysis of the relation between traffic variables, congestion measures and road network structure are provided in Appendix~\ref{app:analysis-traffic-congestion}. Overall, this analysis serves as an example how widely studied traffic questions can be investigated in a data-driven and reproducible setting at a city scale using the introduced \texttt{city-traffic} datasets.

\section{Scalable graph-based traffic forecasting}
\label{sec:traffic-forecasting}

Among various problems in traffic modeling, traffic forecasting, the task of predicting future traffic conditions (e.g., speed and volume) based on historical observations, has received the most significant attention from the machine learning community. Spatiotemporal GNNs have emerged as the dominant approach due to their inherent ability to model complex spatial and temporal dependencies simultaneously \citep{jiang2022graph, cini2023graph}. In this section, we conduct an empirical study and compare several established spatiotemporal models on the proposed \texttt{city-traffic} datasets. Our findings highlight significant scalability limitations in these architectures. To address them, we introduce a simple but effective model that not only successfully scales to these large datasets but also provides forecasting accuracy comparable to the considered methods. The necessary background on spatiotemporal GNNs is provided in Appendix \ref{app:background-spatiotemporal-models}.

\subsection{Scalable approach for spatiotemporal GNNs}

The introduced \texttt{city-traffic} datasets are much larger than those currently used in the literature, so they pose a significant scalability challenge for deep learning models. Considering the models available in Torch Spatiotemporal \citep{cini2022torch} and in the codebase of \texttt{LargeST} \citep{liu2023largest}, the largest previous traffic forecasting dataset, we found that most of them failed to scale on \texttt{city-traffic-M} using a GPU with 80GB VRAM, and even those methods that could carry out training appeared to be very inefficient. Upon closer examination of their architecture it becomes clear that they typically use recurrence, convolution, or attention to process the temporal dimension of data. However, these mechanisms need to maintain a separate vector representation for each timestamp in the lookback window for each node in the graph, which requires large memory and computational resources. In standard, graph-agnostic time series forecasting, several works \citep{oreshkin2019n, zeng2023transformers, zhang2022less, das2023long, li2023revisiting, yi2024frequency} have shown that this issue can be resolved by concatenating all the past time series values into a single input vector and directly mapping it to a single latent representation.

In this work, we propose to exploit this approach for graph-based traffic forecasting. Specifically, we take the idea of encoding each time series in a multivariate dataset into a single vector representation (SVR) with a linear layer and adapt it to graph-based forecasting setting by replacing the following MLP with a GNN that allows representations of different time series to interact through the graph connectivity. In practice, we use a standard mean aggregation common for modern message-passing networks \citep{hamilton2017inductive} and refer to this model as SVR-GNN. Following \citet{platonov2023critical, GraphLand}, we augment our model with skip-connections \citep{he2016deep}, layer normalization \citep{ba2016layer} and MLP blocks, which often significantly improve their performance (see Appendix \ref{app:architecture} for more details). Our experiments show that such an efficient approach, despite its simplicity, often results in forecasting quality competitive with prior methods.

\subsection{Experimental setup}

Since the models considered in our study differ substantially in computational cost, we use a fixed wall-clock training budget of 12 hours for each learnable model. We tune the learning rate and hidden dimension within this budget, as these hyperparameters had the largest effect in preliminary experiments. This keeps the comparison budget-aware while allowing each architecture to choose its own capacity-efficiency trade-off. Our experimental setup is described in more detail in Appendix~\ref{app:experimental-setup}. Below, we describe the models used in our comparison.

First, to establish reference points for model performance, we evaluate several naive forecasting methods that rely on simple heuristics derived from past traffic values. One of them is \emph{previous}, which predicts the most recently observed value at each road segment. We also consider heuristics that use the daily and weekly periodicity in traffic patterns, which is commonly observed in urban traffic dynamics. Namely, we predict traffic variable by using the corresponding value either one day or one week ago from the target timestamp and refer to them as \emph{previous 1 day/week ago}. Next, we include simple statistical baselines such as the global \emph{mean} and \emph{median}, as well as \emph{node-wise mean} and \emph{node-wise median}, which are the mean and median computed independently for each road segment. These naive baselines do not exploit the graph structure. We also evaluate a linear model that can be considered as a basic learnable graph-agnostic method.

Further, we have selected four popular spatiotemporal GNNs from the literature that are frequently used by other works on graph-based time series forecasting and that could scale to our datasets. To process the temporal dimension, these models utilize either recurrence or convolutions: DCRNN \citep{li2018diffusion} is a diffusion convolutional recurrent neural network that exploits recurrent cells supplied with a graph convolution operation; STGCN \citep{yu2018spatio} is a spatiotemporal graph neural network that is composed of alternating temporal and graph convolution operations; GWN \citep{wu2019graph} is a spatiotemporal graph neural network that stacks graph convolutions and causal dilated temporal convolutions; GRUGCN \citep{gao2022equivalence} is a combination of a recurrent temporal encoder and a graph convolutional spatial encoder, which are stacked consecutively. For our experiments, we adapt the implementations from the \texttt{LargeST} repository \citep{liu2023largest}. We also evaluate the proposed scalable traffic forecasting model SVR-GNN.

\begin{table}[t!]
\centering
\caption{Traffic forecasting performance across \texttt{city-traffic} datasets. MAE is reported on the test set. \texttt{OOM} indicates failure to scale on a single A100 80GB GPU with batch size 1.}
\vspace{2pt}
\label{tab:model-performance}
\begin{tabular}{clcccc}
\toprule
&&
\multicolumn{2}{c}{\texttt{city-traffic-M}} &
\multicolumn{2}{c}{\texttt{city-traffic-L}} \\[2pt]
&&
Volume & Speed &
Volume & Speed \\
\midrule
\multirow{7}{*}{\rotatebox{90}{\footnotesize naive baselines}} 
& mean & $2.848$ & $11.704$ & $9.413$ & $11.828$ \\
& median & $2.063$ & $11.161$ & $7.577$ & $11.551$ \\
& node-wise mean & $1.527$ & $5.448$ & $5.355$ & $5.912$ \\
& node-wise median & $1.491$ & $5.375$ & $5.297$ & $5.818$ \\
& previous & $0.957$ & $4.240$ & $2.641$ & $4.576$ \\
& previous 1 day ago & $0.988$ & $5.550$ & $2.808$ & $5.827$ \\
& previous 1 week ago & $0.926$ & $5.476$ & $2.540$ & $5.700$ \\
\midrule
& Linear model & $0.806 \pm 0.000$ & $3.951 \pm 0.001$ & $2.284 \pm 0.000$ & $4.229 \pm 0.001$ \\
\midrule
\multirow{5}{*}{\rotatebox{90}{\footnotesize spatiotemporal}} 
& DCRNN 
& $0.818 \pm 0.008$ 
& $3.494 \pm 0.028$ 
& $2.148 \pm 0.020$ 
& $3.829 \pm 0.013$ \\

& GRUGCN 
& $0.699 \pm 0.001$ 
& $3.136 \pm 0.003$ 
& $1.938 \pm 0.003$ 
& $3.638 \pm 0.008$ \\

& STGCN 
& $0.724 \pm 0.004$ 
& $3.722 \pm 0.084$ 
& \texttt{OOM} 
& \texttt{OOM} \\

& GWN 
& $0.723 \pm 0.005$ 
& $3.523 \pm 0.014$ 
& $2.078 \pm 0.027$ 
& $3.944 \pm 0.063$ \\

& SVR-GNN 
& \textcolor{blue}{$0.673 \pm 0.000$} 
& \textcolor{blue}{$3.098 \pm 0.002$} 
& \textcolor{blue}{$1.847 \pm 0.001$} 
& \textcolor{blue}{$3.454 \pm 0.006$} \\
\bottomrule
\end{tabular}
\vspace{-15pt}
\end{table}

\begin{table}[t!]
\centering
\caption{Inference time (in minutes) across \texttt{city-traffic} datasets.}
\vspace{2pt}
\label{tab:model-efficiency}
\begin{tabular}{lrr}
\toprule
& \texttt{city-traffic-M} & \texttt{city-traffic-L} \\
\midrule
Linear            & $\phantom{.}1.28$ & $\phantom{.}4.08$ \\
DCRNN             & $122.70$           & $176.58$ \\
GRUGCN            & $\phantom{.}17.13$ & $26.50$ \\
STGCN             & $60.15$            & $\texttt{OOM}$ \\
GWN               & $25.00$            & $51.00$ \\
SVR-GNN           & $\phantom{.}2.90$  & $\phantom{.}6.33$ \\
\bottomrule
\end{tabular}
\vspace{-15pt}
\end{table}

\subsection{Experimental results}

First, we compare the forecasting performance of the considered models. The results are shown in Table~\ref{tab:model-performance}. Among the considered naive baselines, the best results for traffic volume prediction are achieved by the predictor taking the value from one week ago from the target timestamp; for speed prediction, the best naive predictor employs the latest known value. These metric values should serve as a necessary sanity check to ensure that the designed models actually capture useful information for the given forecasting task. Thus, as expected, the linear model consistently outperforms the presented naive baselines, which demonstrates that using historical observations is essential for precise traffic forecasting. More advanced spatiotemporal methods, in turn, generally have better performance than graph-agnostic approaches, which indicates that using structural information about the road network is important for accurate traffic forecasting. At the same time, some existing spatiotemporal models face scalability issues on larger city-traffic datasets: for example, STGCN runs out of memory on \texttt{city-traffic-L}. The proposed SVR-GNN achieves the best performance across all reported datasets and target variables, outperforming both simple baselines and more complex spatiotemporal architectures. These results suggest that even a simple non-sequential graph-based aggregation mechanism can effectively capture useful structural information in large-scale road networks, making it a strong and scalable backbone for spatiotemporal traffic forecasting.

Next, we report inference time across \texttt{city-traffic} datasets in Table~\ref{tab:model-efficiency}.
The considered sequential spatiotemporal models, especially DCRNN, GWN, and STGCN, exhibit substantially worse scaling behavior: they require tens or even hundreds of minutes for inference. These scalability issues are caused by the need to maintain and process an explicit temporal state for each input timestamp, which grows linearly with lookback size. In contrast, the proposed SVR-GNN achieves inference time comparable to the linear model, while providing results competitive with other graph-based approaches. This demonstrates that such a non-sequential full-batch design is significantly more scalable and computationally efficient. We hope that our findings will encourage further development of efficient methods for traffic modeling and graph-based spatiotemporal forecasting.

\section{Limitations}
\label{app:limitations}

First, as mentioned in Appendix \ref{app:datasets-details}, an important limitation of our proposed \texttt{city-traffic} datasets is that all traffic measurements are derived from the GPS signal transmitted by drivers who use a specific navigation service. However, we note that this data collection approach reflects a real industrial setting: a production-grade traffic forecasting and urban navigation service that has provided us with the released traffic datasets operates on precisely this type of GPS-derived data to serve tens of millions of users every day, and this is often a vast majority of city residents and drivers. The fact that such systems achieve reliable real-time performance demonstrates that GPS-based measurements, while underestimating absolute traffic volume, preserve the underlying traffic dynamics sufficiently well for practical applications.

We also emphasize that, despite this limitation, the \texttt{city-traffic} datasets represent a substantial improvement over existing publicly available resources (see Section \ref{sec:existing-datasets} for context), and the GPS-based measurements from a large-scale navigation service offer broad and reliable coverage of the overall traffic dynamics that could be hard to achieve with other sources of traffic data. The precision of the captured dynamics is supported by Figure~\ref{fig:weekly-dynamics}: the weekly traffic dynamics in our datasets are perfectly consistent with common real-world traffic behavior --- pronounced morning and evening rush hours on weekdays manifest as increased volume and decreased speed, while weekends exhibit more gradual and uniform traffic patterns. This provides strong evidence that the GPS-derived measurements properly represent the true underlying traffic dynamics.

Further, as outlined in Section \ref{sec:city-traffic-datasets}, the \texttt{city-traffic} datasets include only 4 month of traffic measurements with a 5-minute granularity. However, we clarify that this choice reflects the operational reality of the production traffic monitoring system from which our data is sourced, rather than an arbitrary limitation or preference. The current temporal coverage is constrained by the technical factors of our data provider, and extending it requires additional computational and regulatory effort that we hope will be possible in the future. Our preprocessing pipeline is fully time-agnostic, making such extensions straightforward as more data becomes available. We also note that, even with such a time span, no other public dataset offers this level of spatial detail for urban traffic, so we encourage other providers to share the traffic data of comparable granularity with the research community.

\section{Discussion and future opportunities}
\label{sec:discussion}

In this work, we focus on a key problem in the study of modern urban traffic modeling: the lack of public datasets that combine real city road networks with fine-grained traffic data. To overcome it, we introduce \texttt{city-traffic} --- two public datasets that combine metropolis-scale road networks with high-frequency speed and volume data from dense GPS signals. We explore the potential of the proposed datasets for analysis, showing how road network structure and dynamic traffic data allow for a deeper investigation of traffic patterns. We also conduct a case study on traffic forecasting to highlight critical scalability issues in existing graph-based neural architectures and propose an efficient model to address them. The datasets presented in our work open up several opportunities for future research beyond forecasting: investigating the effects of road infrastructure on traffic flow to inform urban planning; building accurate urban simulators to test the impact of interventions like road closures; studying how well traffic models generalize across different urban environments. We hope that the proposed \texttt{city-traffic} datasets will serve as a foundational data source, allowing researchers to develop and evaluate models that are not only accurate but also scalable and suitable for real-world urban environments. By making large-scale, realistic traffic data more accessible, we want to support future innovation in urban computing and smart city development.

\section*{Acknowledgments}

We thank Mikhail Seleznyov for helpful discussions and valuable suggestions regarding time series forecasting methods. His input on methodological choices and details of experimental design was beneficial for our work. We also thank Ivan Gorin, Aleksei Istomin, and Alexandr Ruchkin for their help with collecting the data for our datasets.

\bibliography{references}
\bibliographystyle{refstyle}


\clearpage
\appendix

\section{Dataset details}
\label{app:datasets-details}

The data used in our benchmark is collected from a widely-used online map and navigation service that estimates traffic congestion and travel time using high-resolution GPS signals transmitted by vehicles. To select the road segments, we take the central geographic point within each city, consider a circular area of a $15$ kilometer radius, and include all road segments located within this area to the dataset. The obtained set of road segments includes the city itself and may also cover some nearby roads.

The traffic volume is estimated based on the number of vehicles that traverse each road segment during a specific timestamp interval, as inferred from aggregated GPS traces. It is important to note that the number of traverses represents an estimate rather than the actual traffic flow, as it is derived solely from vehicles equipped with GPS. Consequently, the reported values can underestimate the true traffic volume, but they cover the major part of this signal and adequately represent the actual dynamics of the traffic volume. The speed estimation is also derived from these GPS signals, using a proprietary internal algorithm developed by the service of data provider.

Some characteristics of our datasets are reported in Table~\ref{tab:datasets-characteristics}.

\begin{table*}[h]
\centering
    \caption{Characteristics of new \texttt{city-traffic} datasets.}
\label{tab:datasets-characteristics}
\begin{center}
\begin{tabular}{lcc}
\toprule
                                 & \texttt{city-traffic-M}   & \texttt{city-traffic-L}   \\
\midrule
\# nodes                         & $53{,}530$                & $94{,}009$                \\
\# edges                         & $121{,}236$               & $164{,}424$               \\
is directed                                & \cmark               & \cmark                 \\
\midrule
\# timestamps                    & $35{,}449$                & $35{,}449$                \\
\# train timestamps              & $26{,}208$                & $26{,}208$                \\
\# validation timestamps                & $4{,}032$                 & $4{,}032$                 \\
\# test timestamps               & $5{,}209$                 & $5{,}209$                 \\
\midrule
train start                             & \texttt{Jul \phantom{0}1st 2024 00:00} & \texttt{Jul \phantom{0}1st 2024 00:00} \\
validation start                               & \texttt{Sep 30th 2024 00:00} & \texttt{Sep 30th 2024 00:00} \\
test start                              & \texttt{Oct 14th 2024 00:00}   & \texttt{Oct 14th 2024 00:00}   \\
test end                                & \texttt{Nov \phantom{0}1st 2024 02:00} & \texttt{Nov \phantom{0}1st 2024 02:00}    \\
\midrule
avg. in-degree  / avg. out-degree                     & $2.264$              & $1.749$                \\
avg. node degree (undirected)          & $3.652$              & $2.970$                \\
Gini coefficient of degree distribution & $0.9$                & $0.9$                  \\
\bottomrule
\end{tabular}
\end{center}
\end{table*}

Each node in the dataset represents an individual road segment and has a set of 26  attributes, including categorical and binary indicators of road type, accessibility, and structural properties. The full list of feature names is the following:
\begin{itemize}[leftmargin=20pt]
    \item \texttt{category} --- functional category of the road segment (e.g., major arterial, residential, service);
    \item \texttt{edge\_type} --- encodes the type of connection between the road segments;
    \item \texttt{speed\_mode} --- type of speed regulation pattern allowed on the segment (e.g., high-speed corridor, restricted-speed street);
    \item \texttt{speed\_limit} --- the maximum legal speed limit on the segment;
    \item \texttt{region\_id} --- identifier of the administrative or city district containing the segment;
    \item \texttt{can\_bind\_to\_reverse\_edge} --- indicates whether the segment allows binding to a reverse-direction edge;
    \item \texttt{dismount\_bike} --- indicates if cyclists are required to dismount on the segment;
    \item \texttt{has\_masstransit\_lane} --- indicates if the segment has a dedicated lane for public or mass transit;
    \item \texttt{ends\_with\_crosswalk} --- indicates if the segment ends with a pedestrian crosswalk;
    \item \texttt{ends\_with\_toll\_post} --- indicates if the segment ends with a toll post;
    \item \texttt{is\_in\_poor\_condition} --- indicates whether the road surface is in poor condition;
    \item \texttt{is\_paved} --- indicates whether the segment is paved;
    \item \texttt{is\_restricted\_for\_trucks} --- indicates whether the segment is restricted for trucks;
    \item \texttt{is\_toll} --- indicates whether the segment is a toll road;
    \item \texttt{access\_[0...5]}\footnote{There is a separate feature for each of 6 masks.} --- boolean masks for road accessibility by different undisclosed transport modes (exact mapping to vehicle types will be released by the provider);
    \item \texttt{length} --- length of the road segment (in meters);
    \item \texttt{num\_segments} --- number of consecutive sub-segments composing the road segment;
    \item \texttt{x\_coordinate\_start} --- latitude of the segment’s start point;
    \item \texttt{y\_coordinate\_start} --- longitude of the segment’s start point;
    \item \texttt{x\_coordinate\_end} --- latitude of the segment’s end point;
    \item \texttt{y\_coordinate\_end} --- longitude of the segment’s end point.
\end{itemize}

Note that \texttt{speed\_limit} feature represents an ordinal encoding of the speed limit value set on the road segment, and this does not apply to the traffic speed itself that is treated as continuous target variable in forecasting task. Thus, we provide the correspondence of particular feature values and their ordinal codes:
\begin{itemize}[leftmargin=20pt]
    \item $\texttt{NaN} \to 0$;
    \item $5\,km/h \to 1$;
    \item $20\,km/h \to 2$;
    \item $30\,km/h \to 3$;
    \item $40\,km/h \to 4$;
    \item $50\,km/h \to 5$;
    \item $60\,km/h \to 6$;
    \item $70\,km/h \to 7$;
    \item $80\,km/h \to 8$;
    \item $90\,km/h \to 9$;
    \item $100\,km/h \to 10$;
    \item $110\,km/h \to 11$;
\end{itemize}

\clearpage

\section{Background on spatiotemporal graph neural networks}
\label{app:background-spatiotemporal-models}

One of the pioneering spatiotemporal forecasting models is DCRNN~\citep{li2018diffusion}, which effectively combined graph diffusion convolutions with recurrent neural networks (RNNs) to model the spatial and temporal components. Further, STGCN \citep{yu2018spatio} replaced RNNs with temporal convolutional layers, thus achieving better computational efficiency and introducing a popular architectural pattern of alternating between spatial and temporal processing modules. Later research focused on more intricate methods to improve predictive accuracy. For instance, GWN \citep{wu2019graph} and AGCRN \citep{bai2020adaptive} proposed the use of adaptive adjacency matrices to learn latent spatial dependencies directly from data, freeing models from relying on a predefined graph structure. At the same time, a significant trend has been the integration of the attention mechanism: models like ASTGCN \citep{guo2019attention} and GMAN \citep{zheng2020gman} were designed to dynamically weigh the importance of different nodes and time steps, aiming to capture more complex spatiotemporal patterns.

Together with other examples, these works mark a notable trend towards increasing architectural complexity. Most recent methods incorporate multiple sophisticated components, such as hierarchical attention or adaptive graph learning, which can introduce significant computational overhead. While these models have demonstrated strong performance on existing benchmarks, their development has been shaped by the very limitations of those datasets, including small graphs with a few hundred nodes and simple graph topology, which have de-emphasized computational efficiency. Consequently, it remains a critical and unaddressed question whether these established architectures can be used on metropolis-scale road networks with tens of thousands of nodes that characterize real urban settings. Our work investigates this question through experiments in Section \ref{sec:traffic-forecasting}.

\section{Distribution of some road segment properties}
\label{app:dist-segment-properties}

\begin{figure}[h!]
    \centering
    \includegraphics[width=0.9\linewidth]{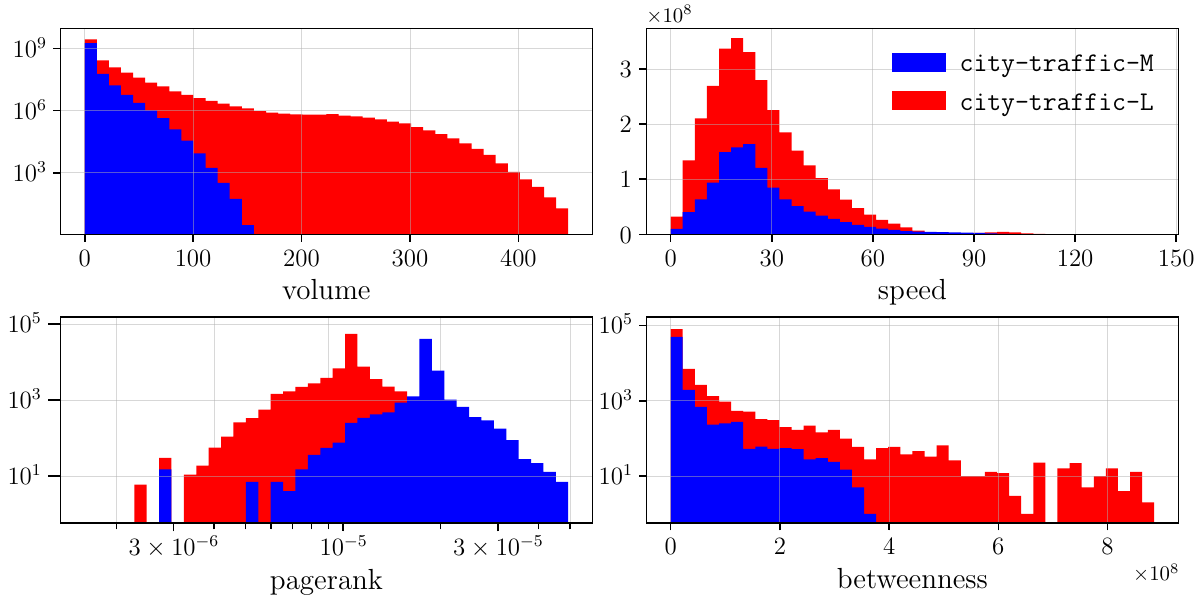}
    \caption{Histograms of traffic variables (top row), and directed PageRank and betweenness (bottom row) across all roads segments in the proposed datasets.}
    \label{fig:joint-hist}
    \vspace{-10pt}
\end{figure}

Figure~\ref{fig:joint-hist} shows the histograms of traffic variables along with the centrality measures in each of the proposed \texttt{city-traffic} datasets. The distribution of traffic volume is strongly skewed, and both variables have substantial spread, reflecting heterogeneity across road segments and time. The amount of traffic also varies significantly across the considered cities, which indicates their different scale and demand. Next, both the considered centrality measures vary substantially across road segments in both datasets. In particular, betweenness centrality is highly heterogeneous and spans multiple orders of magnitude, indicating that the structural roles of road segments are distributed unevenly across the network, consistent with observations reported for urban street graphs in prior work \citep{crucitti2006centrality,kirkley2018betweenness}.

\section{Comparison of \texttt{city-traffic-M} and \texttt{city-traffic-L}}
\label{app:cities-differences}
While both datasets follow the same construction methodology, there are several notable differences between \texttt{city-traffic-M} and \texttt{city-traffic-L} that make them complementary benchmarks.

In terms of scale, \texttt{city-traffic-M} contains $53{,}530$ road segments and $121{,}236$ directed edges, while  \texttt{city-traffic-L} is almost twice as large, with $94{,}009$ segments and $164{,}424$ edges. The higher spatial resolution of \texttt{city-traffic-L} poses a particular challenge for the scalability of spatiotemporal models, as the number of graph nodes directly determines memory and runtime costs.

In terms of topological properties, the two cities also vary significantly and have a different urban structure. \texttt{city-traffic-L} features a complex structure shaped by a large river crossing the metropolitan area, which has led to the development of multiple islands connected by bridges. This creates bottlenecks and high-traffic corridors that models must capture. By contrast, \texttt{city-traffic-M} lacks such a riverine structure; its road network is more uniform, with a grid-like arrangement and wide avenues even in the central districts. Average node degree of a road network also differs between the datasets: \texttt{city-traffic-M} has an average undirected degree of $3.65$, while \texttt{city-traffic-L}’s average is $2.97$. This reflects the higher density and branching structure of the smaller city versus the sparser but more geographically constrained connectivity of the larger one.

While the average traffic speed values are comparable between the two datasets, the same statistic for traffic volume differs significantly: \texttt{city-traffic-L} records substantially higher overall volume, which reflects its larger scale. The weekly dynamics, shown in Figure \ref{fig:weekly-dynamics}, indicate more pronounced rush-hour congestion patterns in \texttt{city-traffic-L}.

\begin{figure}[h]
    \centering
    \includegraphics[width=0.9\linewidth]{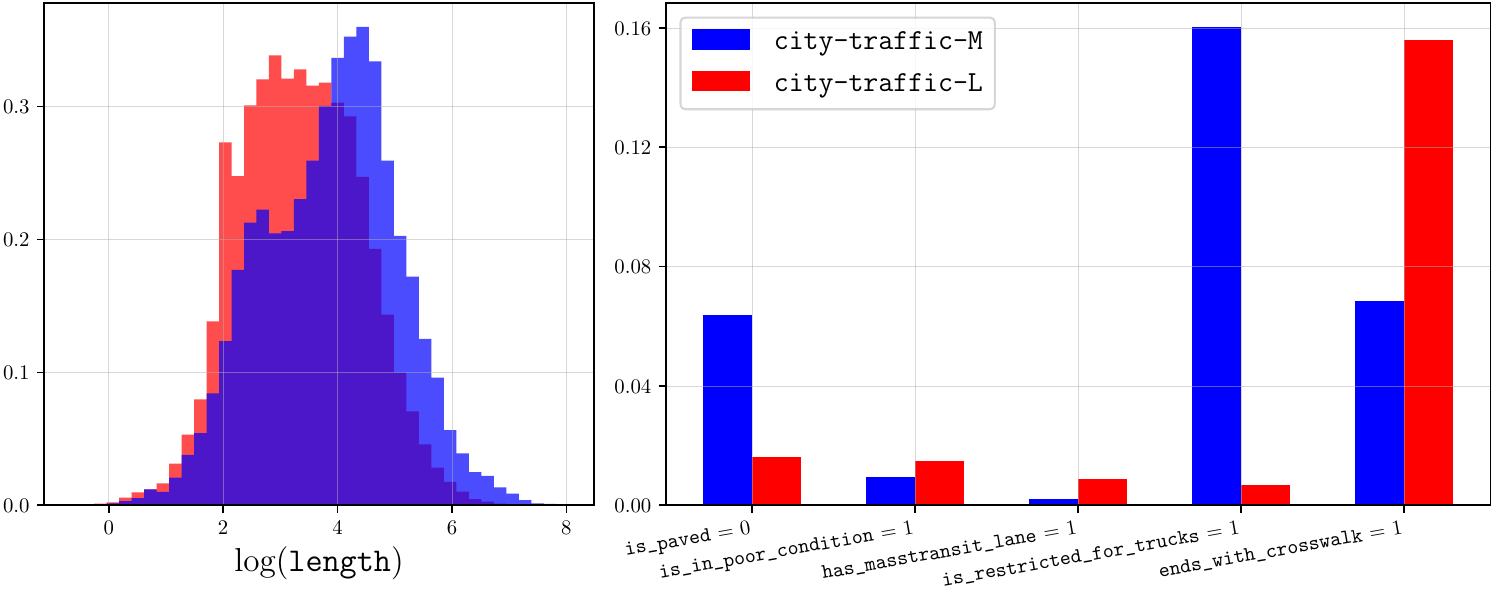}
    \caption{Distribution of some spatial features in the proposed datasets.}
    \label{fig:spatial-features}
    \vspace{-10pt}
\end{figure}

Both datasets provide the same 26 static attributes per road segment. However, their distribution is different for the two proposed datasets. As Figure \ref{fig:spatial-features} shows, \texttt{city-traffic-L} has a greater fraction of paved roads, and there are also notably more roads with crosswalks at their endpoints. On the other hand, \texttt{city-traffic-M} has longer continuous road segments on average, and the fraction of roads restricted for trucks is much greater.

Taken together, the two datasets provide complementary perspectives: \texttt{city-traffic-M} highlights fine-grained dynamics in a compact road network, while \texttt{city-traffic-L} captures large-scale, heterogeneous urban traffic with more complex network structure. This difference is essential for developing models that generalize across diverse city types, rather than overfitting to one particular topology or traffic regime.

\section{Further analysis of traffic variables and congestion measures}
\label{app:analysis-traffic-congestion}

\begin{figure}[h!]
\centering
\includegraphics[width=\linewidth]{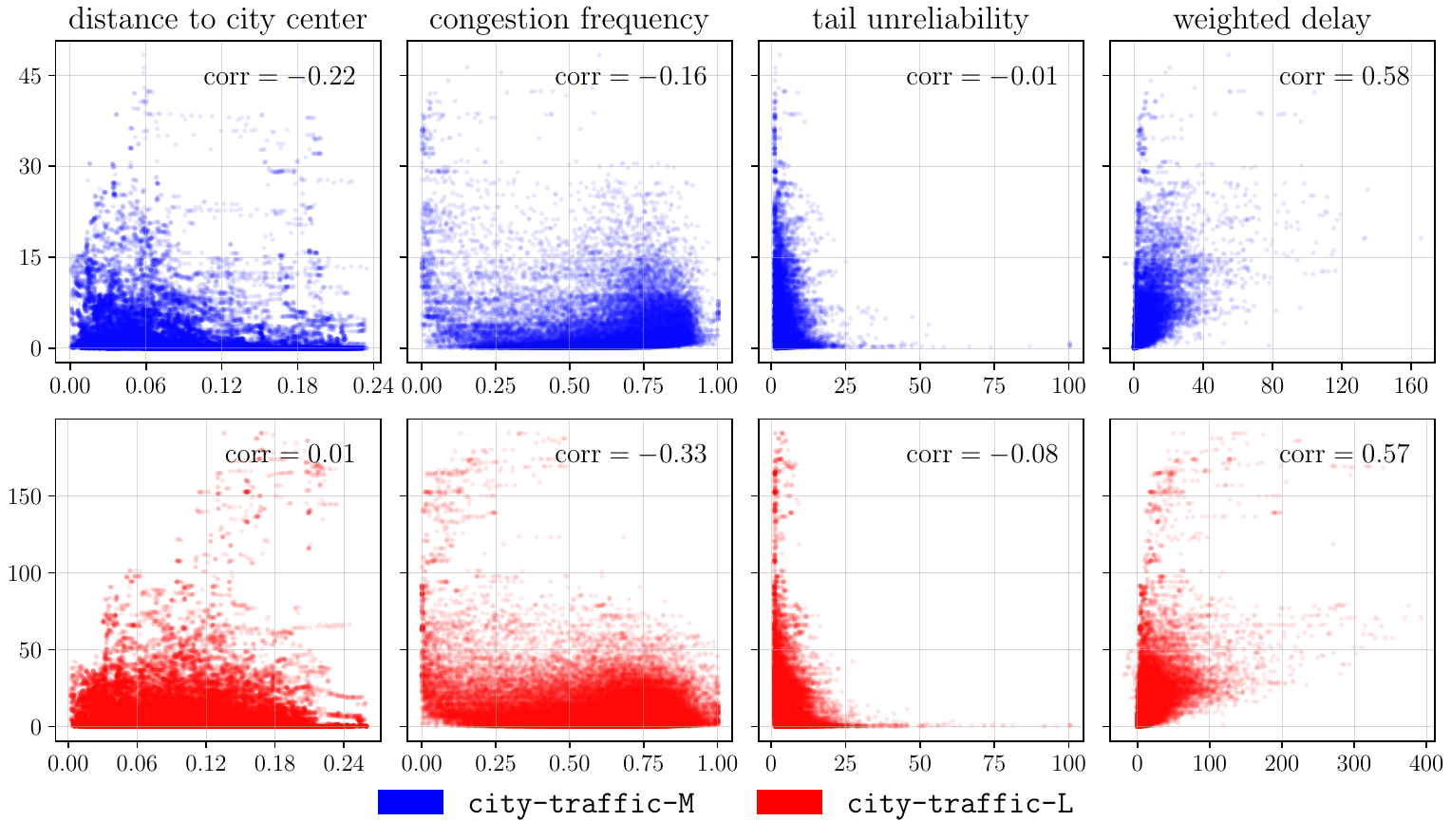}
\caption{Scatter plots of traffic volume (shared $y$-axis) against distance to city center (the leftmost column) and congestion metrics CF, TU and WD (three right columns). Top and bottom rows correspond to \texttt{city-traffic-M} and \texttt{city-traffic-L}, respectively. Each panel is annotated with Pearson correlation coefficient.}
\label{fig:joint-scatter-volume-vs-rest}
\end{figure}

As can be seen on Figure \ref{fig:joint-scatter-volume-vs-rest}, traffic volume has strong positive correlation with WD, primarily because of its definition where it basically integrates the weighted traffic volume. At the same time, it has slight negative correlation with CF, since larger traffic volumes are transferred at higher speed, which prevents congestion events from occurring frequently. Finally, it has almost no correlation with TU, which shows that roads may be reliable or not (which is by definition about outliers, not cumulative effect) regardless of the traffic volume registered on average.

\begin{figure}[h!]
\centering
\includegraphics[width=\linewidth]{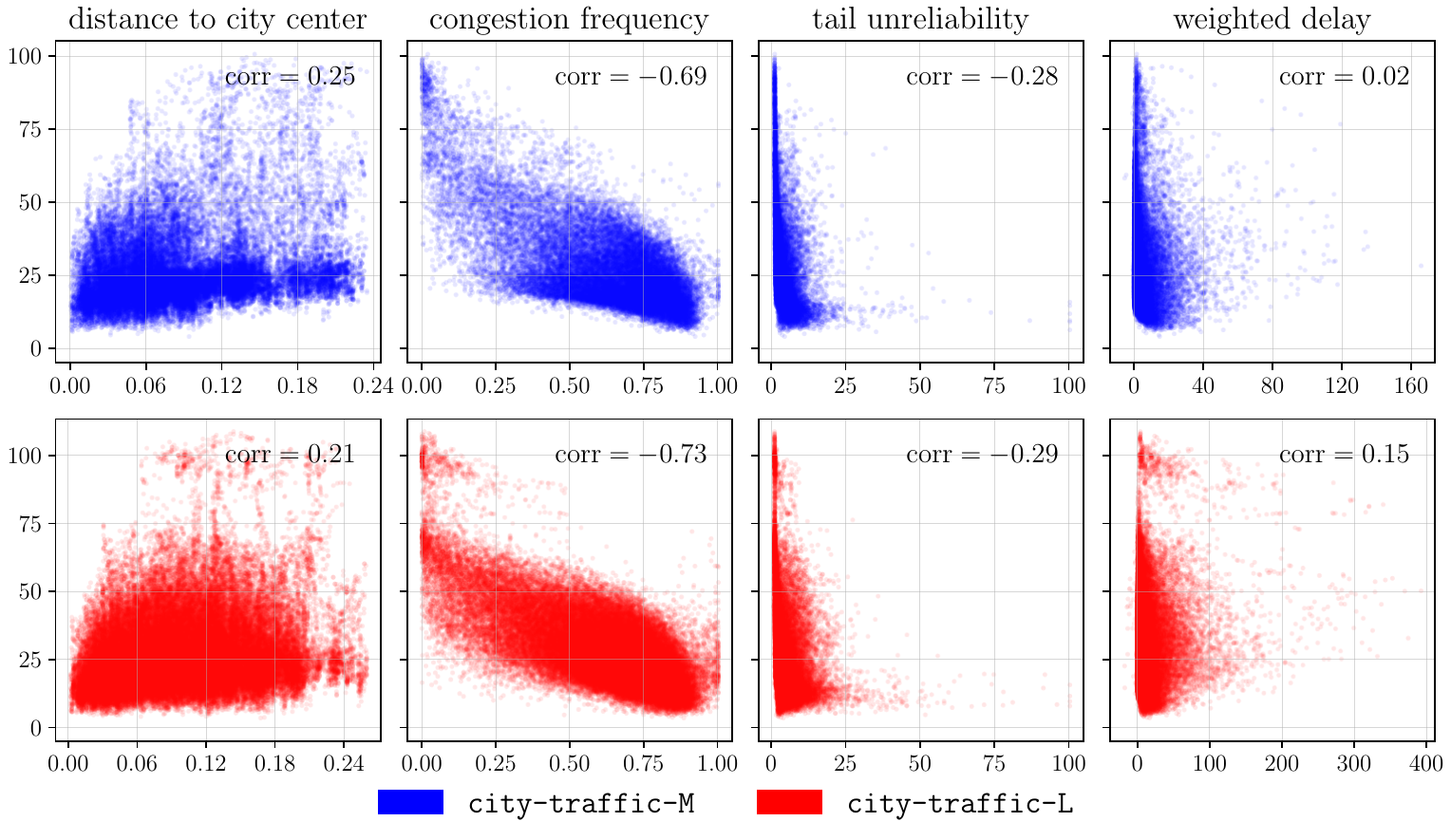}
\caption{Scatter plots of traffic speed (shared $y$-axis) against distance to city center (the leftmost column) and congestion metrics CF, TU and WD (three right columns). Top and bottom rows correspond to \texttt{city-traffic-M} and \texttt{city-traffic-L}, respectively. Each panel is annotated with Pearson correlation coefficient.}
\label{fig:joint-scatter-speed-vs-rest}
\vspace{-10pt}
\end{figure}

Next, Figure \ref{fig:joint-scatter-speed-vs-rest} shows that traffic speed has strong negative correlation with CF, since higher average speed reduces the opportunities to register the speed value much lower than the maximum one. For the same reason, traffic speed is negatively correlated with the TU measure. Regarding distance to the city center, it is positively correlated with traffic speed, which can be explained by the fact that drivers typically can afford to move at higher speed further from the city center, as there are more bypass and highway roads that are intended for fast transfer between distant regions. It is also notably correlated with traffic volume on the smaller \texttt{city-traffic-M}, since its most necessary infrastructure is typically located near the city center, but almost not correlated on the larger \texttt{city-traffic-L}, since important facilities are more widely distributed throughout the city and may be located far from its center.

\section{Visualization of centrality measures in \texttt{city-traffic} road networks}
\label{app:centrality-layouts}

\begin{figure*}[h!]
    \centering
    \begin{subfigure}[t]{0.45\linewidth}
        \centering
        \includegraphics[width=\linewidth]{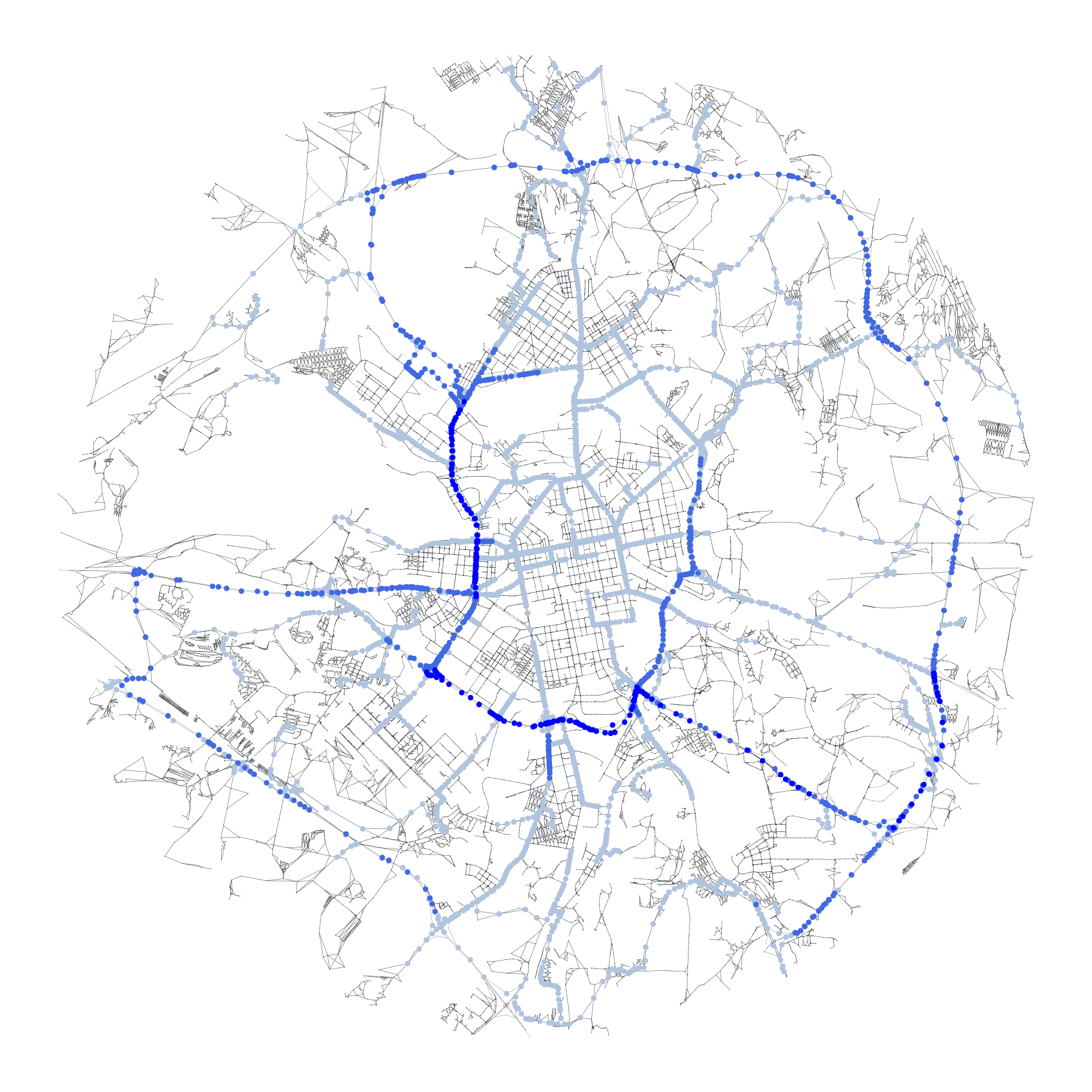}
        \caption{Betweenness values for \textcolor[named]{top200M}{top-$200$}, \textcolor[named]{top1000M}{top-$1000$} and \textcolor[named]{top5000M}{top-$5000$} road segments on \texttt{city-traffic-M}}
        \label{fig:centrality-visualization-betweenness-M}
    \end{subfigure}
    \hspace{0.05\linewidth}
    \begin{subfigure}[t]{0.45\linewidth}
        \centering
        \includegraphics[width=\linewidth]{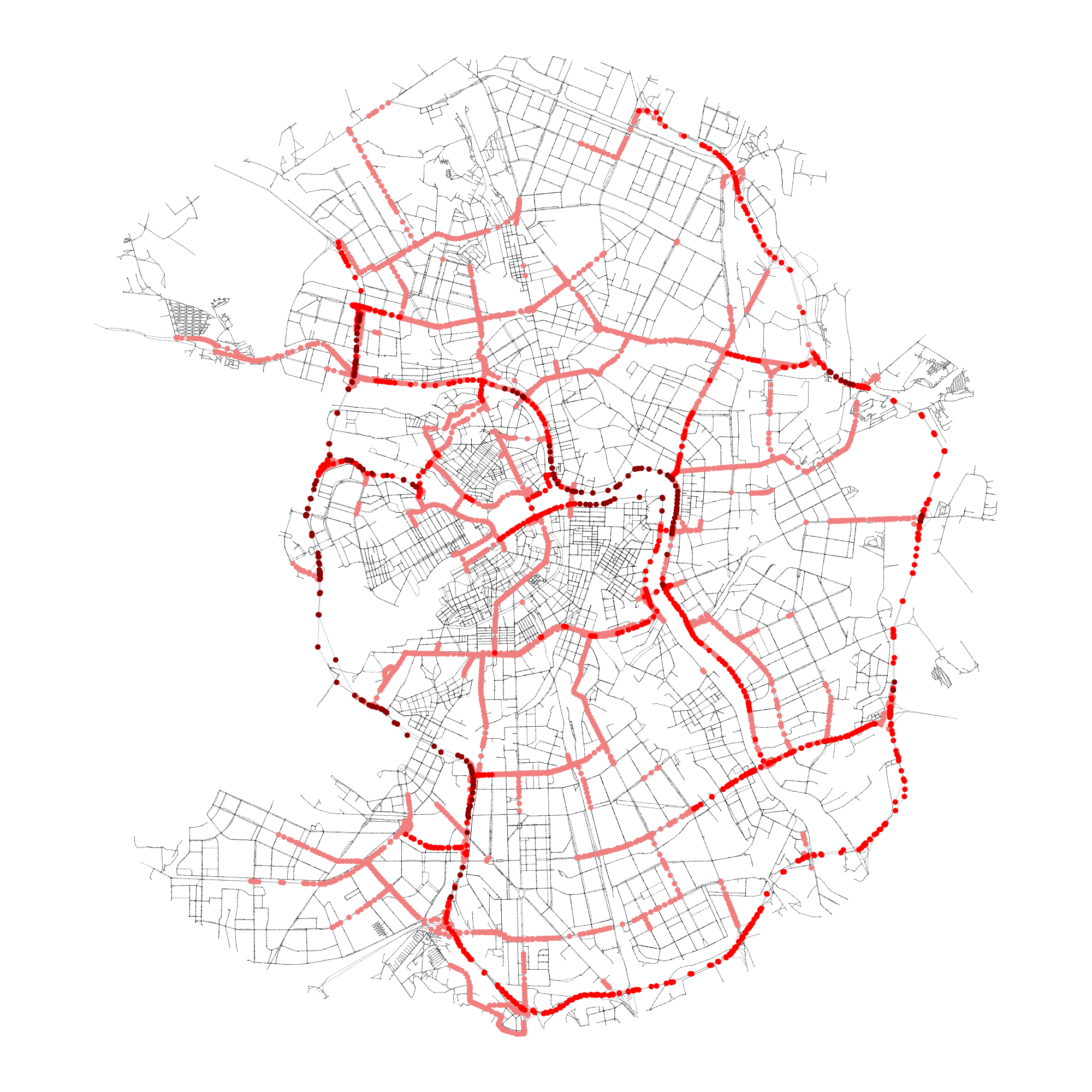}
        \caption{Betweenness values for \textcolor[named]{top200L}{top-$200$}, \textcolor[named]{top1000L}{top-$1000$} and \textcolor[named]{top5000L}{top-$5000$} road segments on \texttt{city-traffic-L}}
        \label{fig:centrality-visualization-betweenness-L}
    \end{subfigure}
    \begin{subfigure}[t]{0.45\linewidth}
        \centering
        \includegraphics[width=\linewidth]{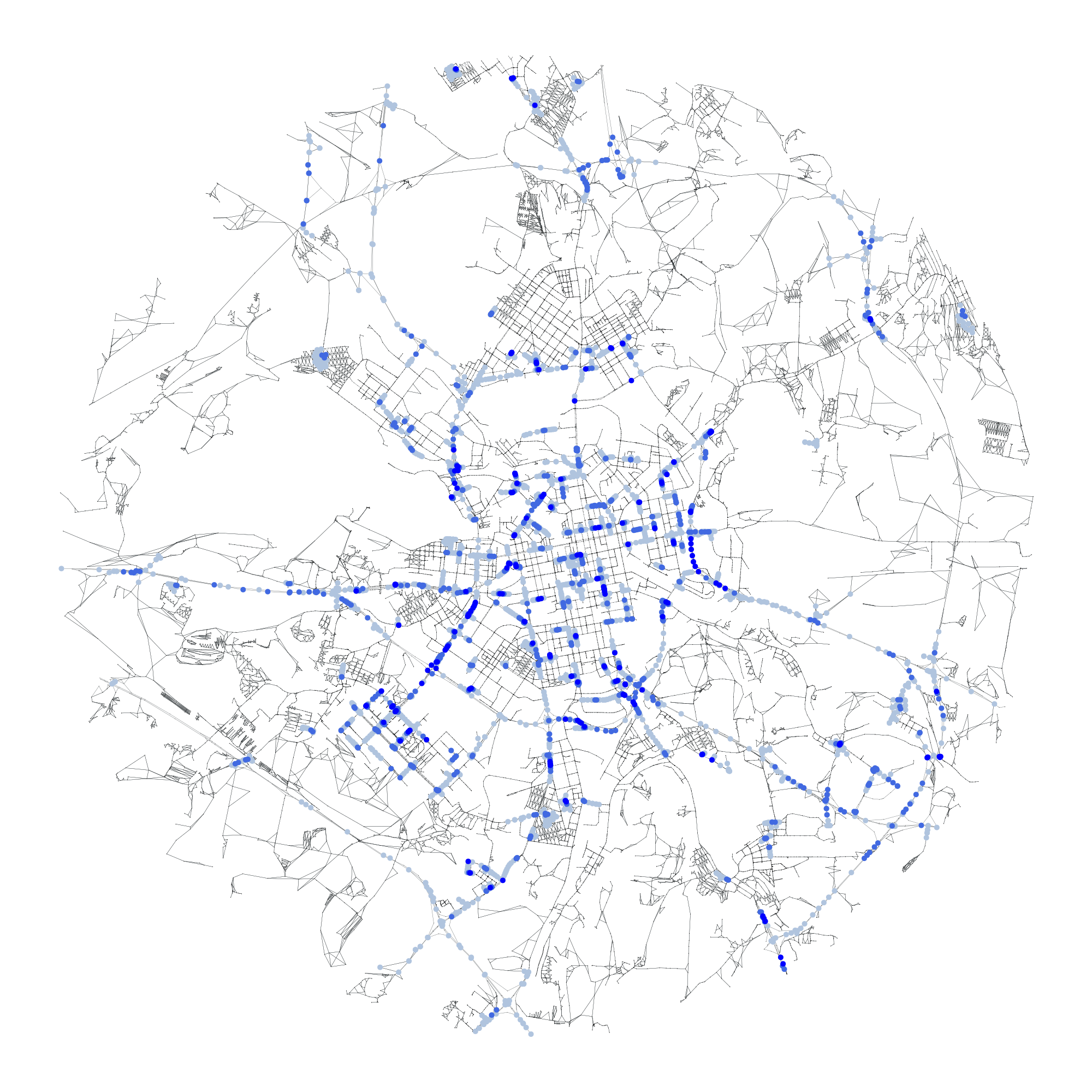}
        \caption{PageRank values for \textcolor[named]{top200M}{top-$200$}, \textcolor[named]{top1000M}{top-$1000$} and \textcolor[named]{top5000M}{top-$5000$} road segments on \texttt{city-traffic-M}.}
        \label{fig:centrality-visualization-pagerank-M}
    \end{subfigure}
    \hspace{0.05\linewidth}
    \begin{subfigure}[t]{0.45\linewidth}
        \centering
        \includegraphics[width=\linewidth]{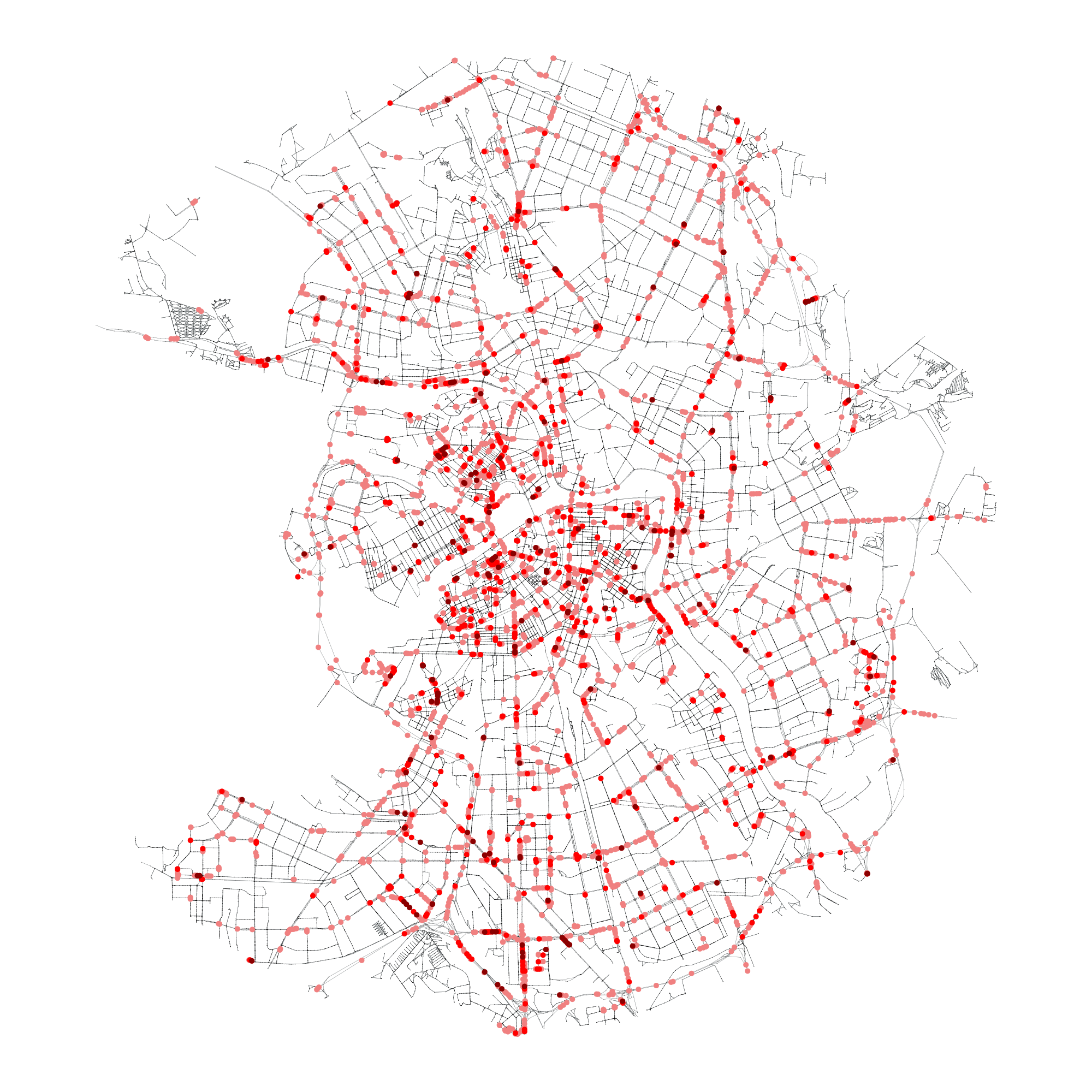}
        \caption{PageRank values for \textcolor[named]{top200L}{top-$200$}, \textcolor[named]{top1000L}{top-$1000$} and \textcolor[named]{top5000L}{top-$5000$} road segments on \texttt{city-traffic-L}}
        \label{fig:centrality-visualization-pagerank-L}
    \end{subfigure}
    \caption{Road segments with the greatest values of centrality measures on the proposed datasets. The top row is for PageRank, the bottom row is for betweenness. The left column is for \texttt{city-traffic-M}, the right column is for \texttt{city-traffic-L}.}
    \label{fig:centrality-visualization}
\end{figure*}

\clearpage

\section{Visualization of congestion measures in \texttt{city-traffic} road networks}
\label{app:congestion-layouts}

\begin{figure*}[h!]
    \centering
    \begin{subfigure}[t]{0.42\linewidth}
        \centering
        \includegraphics[width=\linewidth]{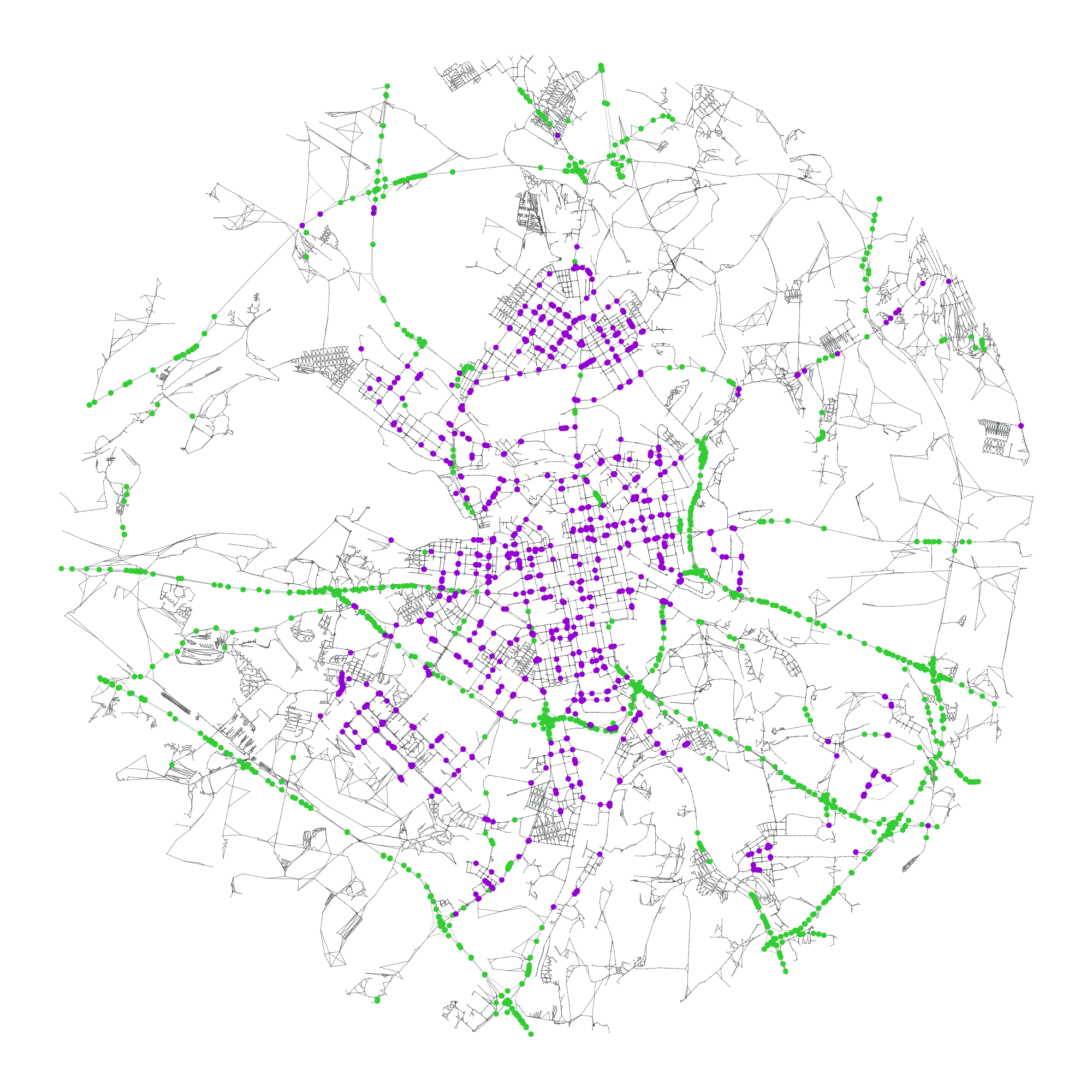}
        \caption{CF on \texttt{city-traffic-M}}\label{fig:CF-M}
    \end{subfigure}
    \hspace{0.05\linewidth}
    \begin{subfigure}[t]{0.42\linewidth}
        \centering
        \includegraphics[width=\linewidth]{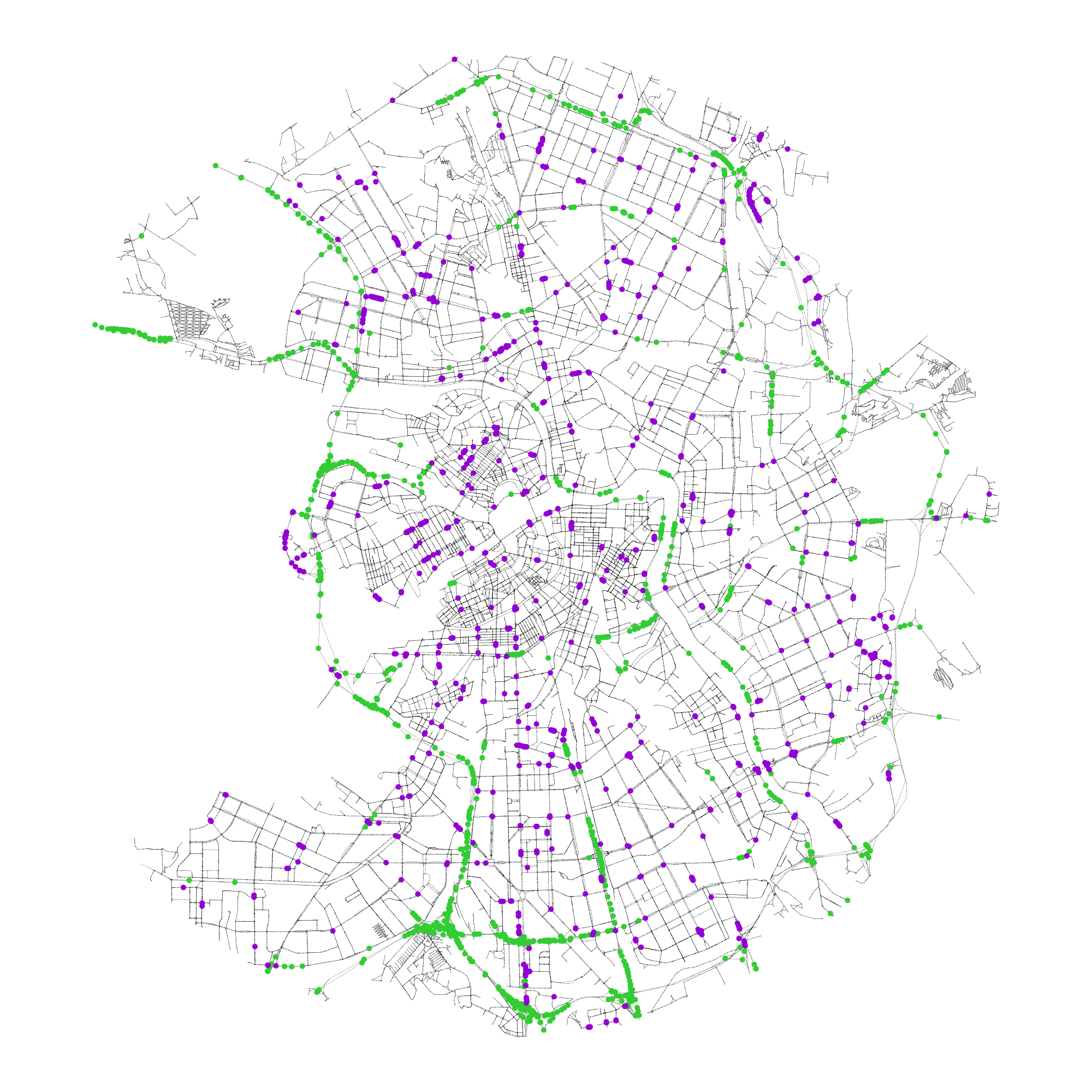}
        \caption{CF on \texttt{city-traffic-L}}\label{fig:CF-L}
    \end{subfigure}

    \begin{subfigure}[t]{0.42\linewidth}
        \centering
        \includegraphics[width=\linewidth]{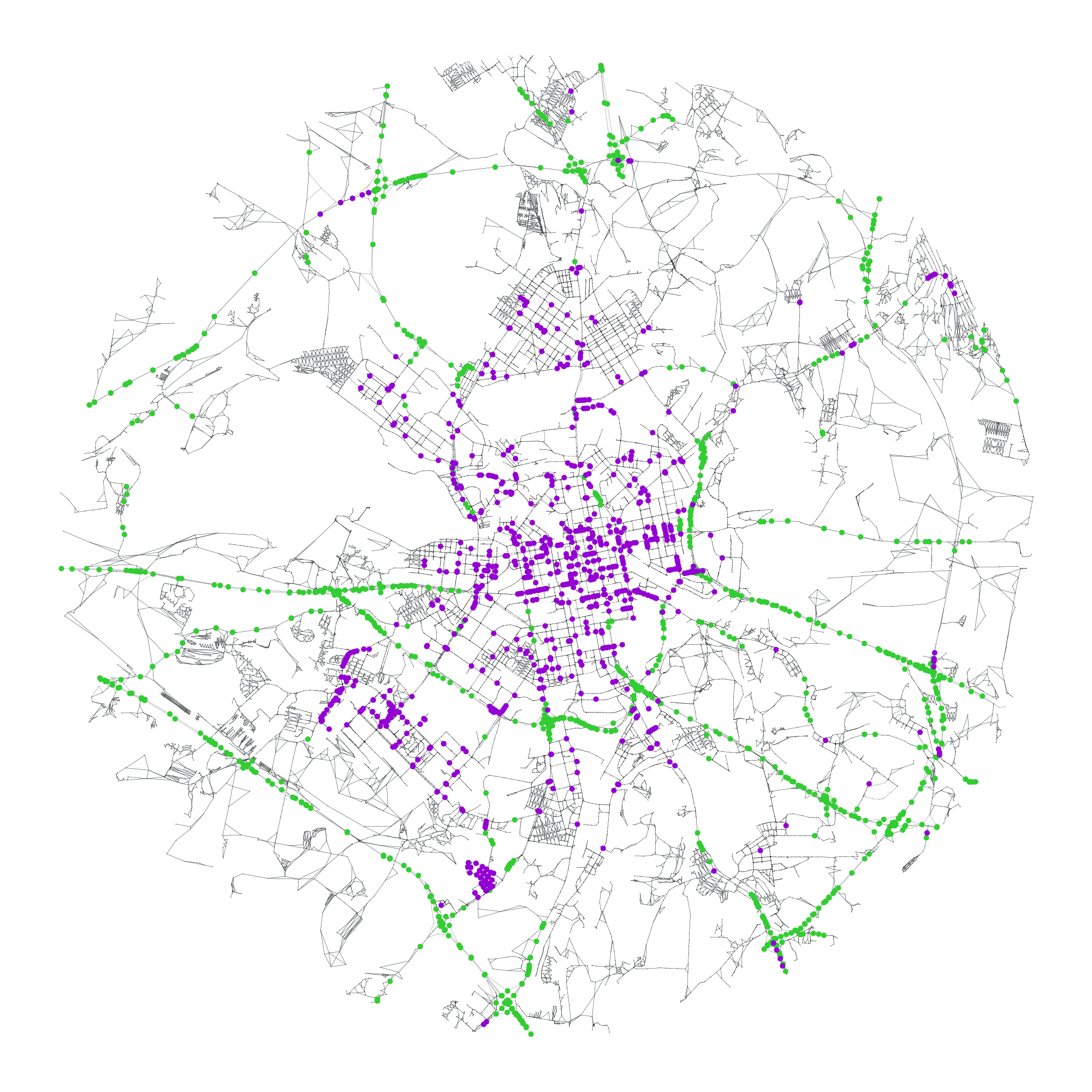}
        \caption{TU on \texttt{city-traffic-M}}\label{fig:TU-M}
    \end{subfigure}
    \hspace{0.05\linewidth}
    \begin{subfigure}[t]{0.42\linewidth}
        \centering
        \includegraphics[width=\linewidth]{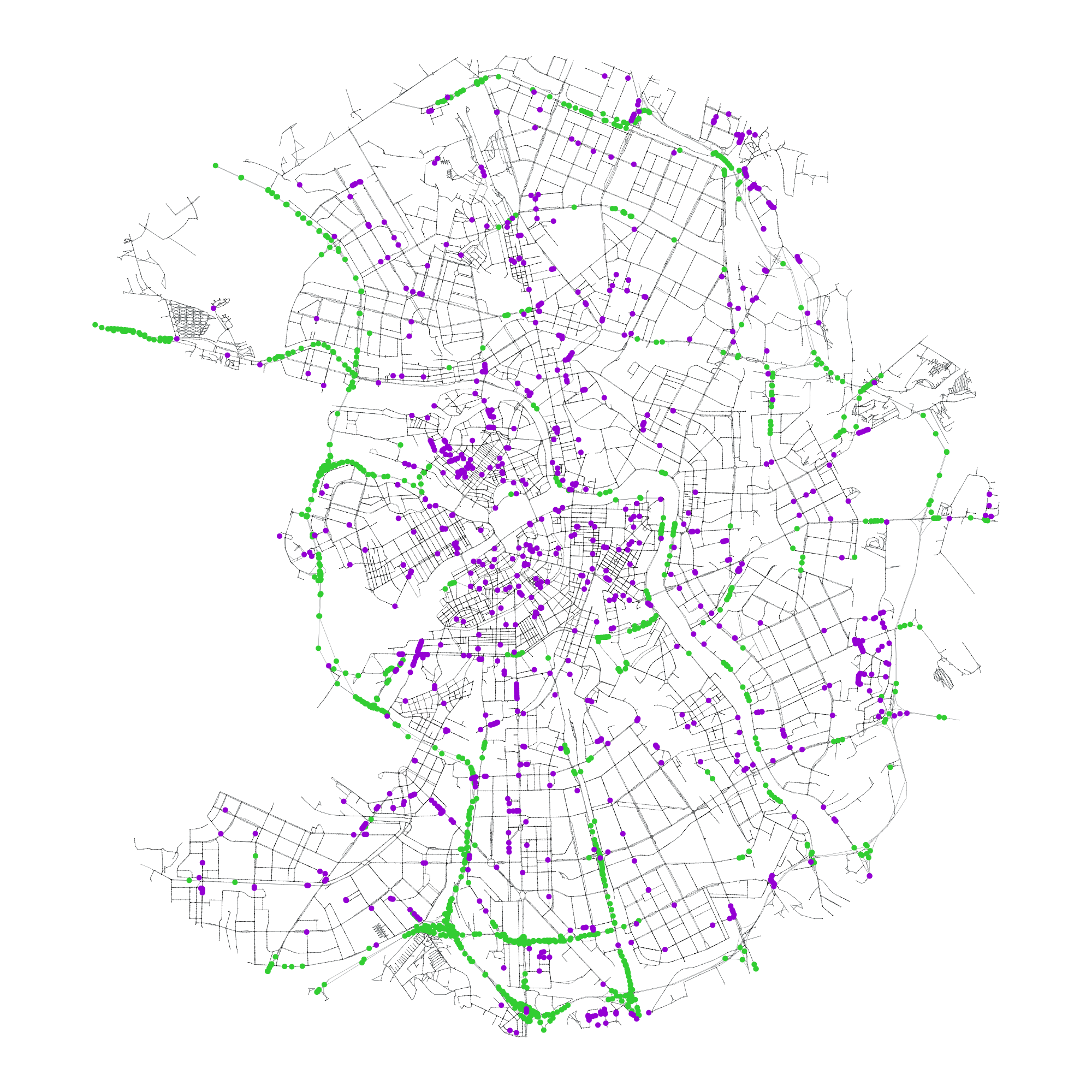}
        \caption{TU on \texttt{city-traffic-L}}\label{fig:TU-L}
    \end{subfigure}

    \begin{subfigure}[t]{0.42\linewidth}
        \centering
        \includegraphics[width=\linewidth]{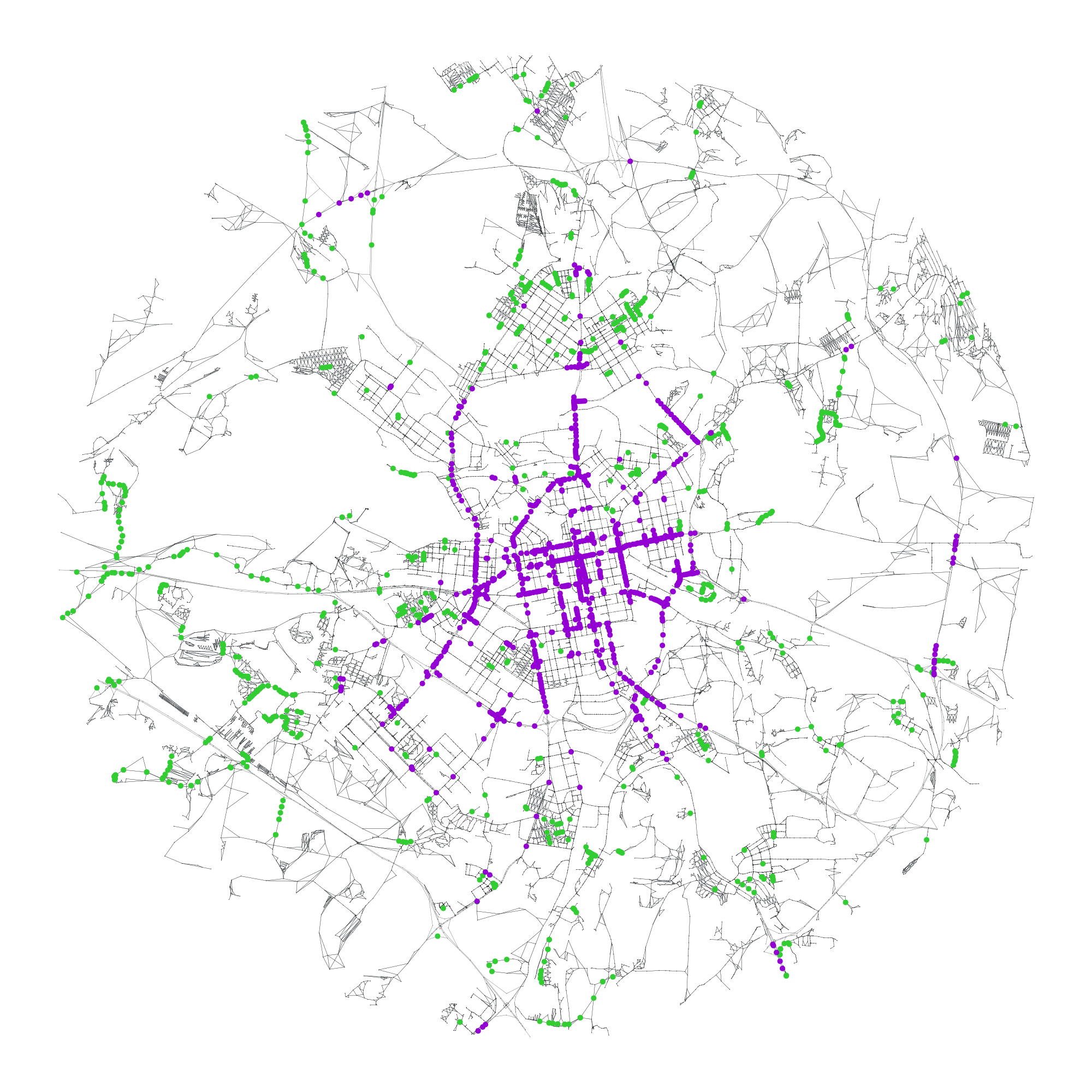}
        \caption{WD on \texttt{city-traffic-M}}\label{fig:WD-M}
    \end{subfigure}
    \hspace{0.05\linewidth}
    \begin{subfigure}[t]{0.42\linewidth}
        \centering
        \includegraphics[width=\linewidth]{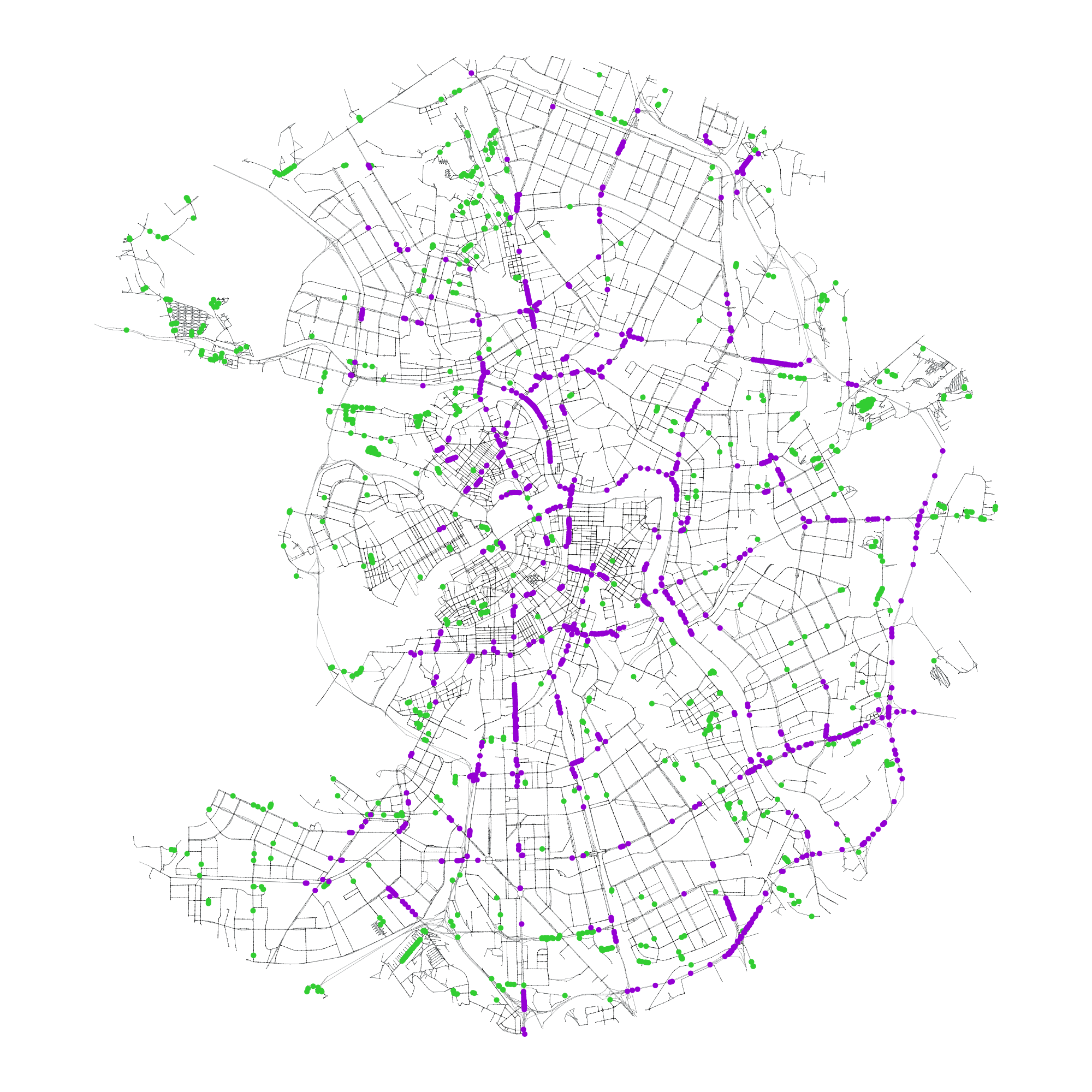}
        \caption{WD on \texttt{city-traffic-L}}\label{fig:WD-L}
    \end{subfigure}
    \caption{Road segments with the \textcolor{top}{greatest} and the \textcolor{bottom}{smallest} values of congestion measures on the proposed datasets (only 1000 roads are taken from both tails of distribution). The first row is for CF, the second row is for TU, the third row is for WD. The left column is for \texttt{city-traffic-M}, the right column is for \texttt{city-traffic-L}.}
    \label{fig:congestion-visualization}
\end{figure*}

\clearpage

\section{Relation between spatial road features and traffic variables}
\label{app:relation}

In this section, we provide several figures with the weekly dynamics of target variables for different road subsets depending on their static attributes and discuss how various spatial road features can affect the traffic volume and speed.

\begin{figure}[h]
    \centering
    \begin{subfigure}{\linewidth}
    \centering
    \includegraphics[width=\linewidth]{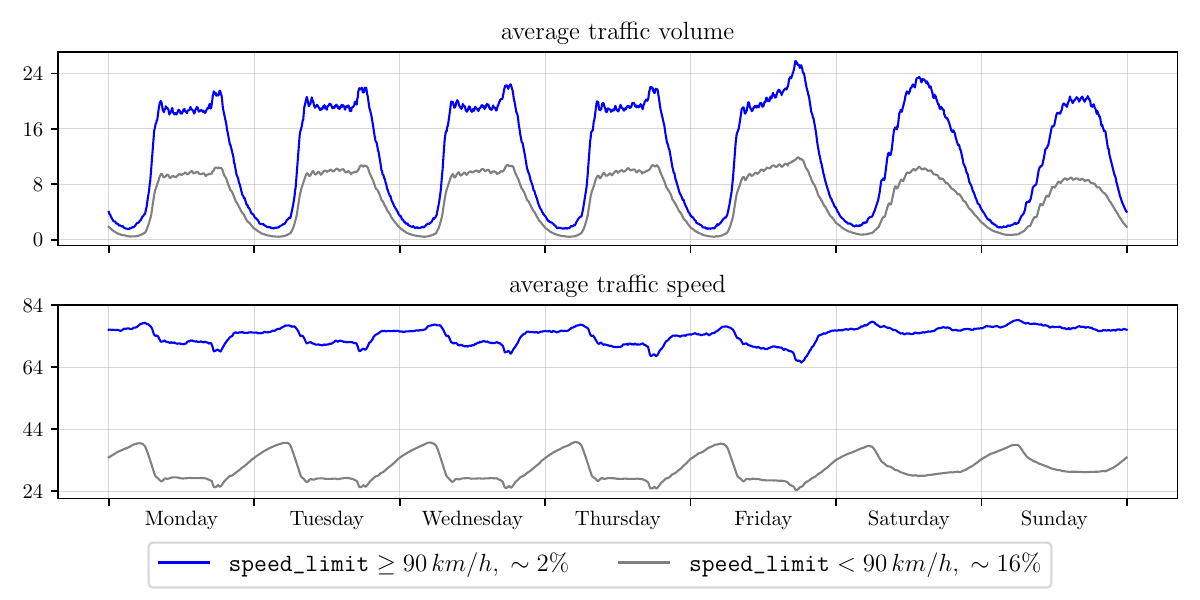}
    \caption{\texttt{city-traffic-M}}
    \end{subfigure}
    \begin{subfigure}{\linewidth}
    \centering
    \includegraphics[width=\linewidth]{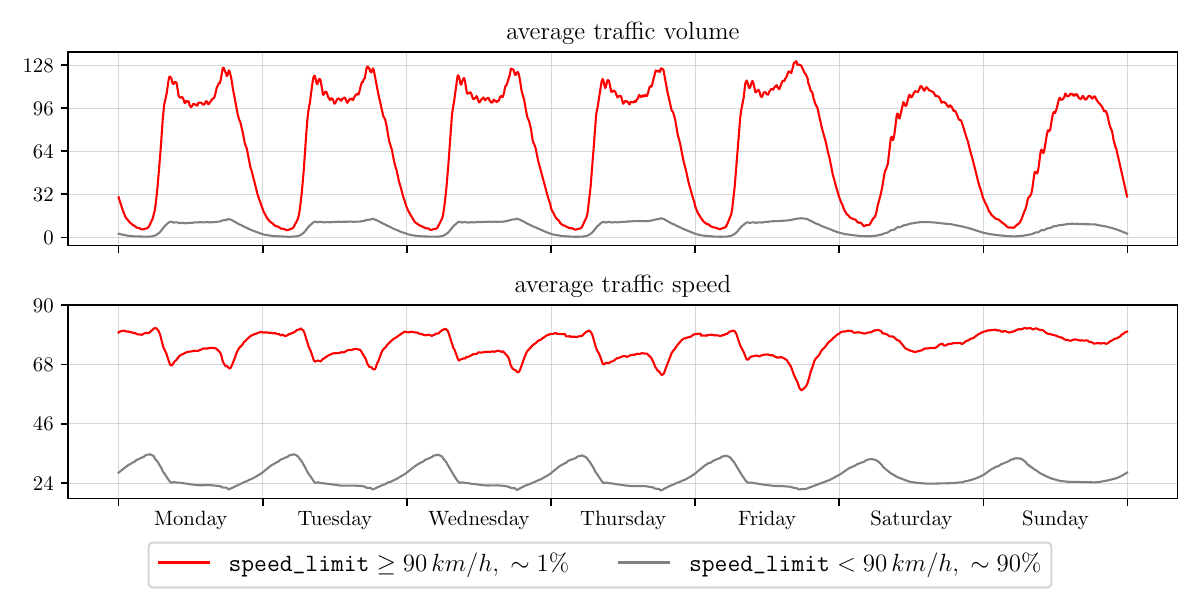}
    \caption{\texttt{city-traffic-L}}
    \end{subfigure}
    \caption{The weekly target dynamics averaged across different road subsets depending on if they have $\texttt{speed\_limit} = 90\,km/h$. The percentage in the legend denotes the fraction of nodes in the corresponding category.}
    \label{fig:targets-dynamics-limit}
    \vspace{-20pt}
\end{figure}

In Figure \ref{fig:targets-dynamics-limit}, we show the dynamics of target variables across the roads with a specific value of the \texttt{speed\_limit} feature (in our case, we use the subset with $\texttt{speed\_limit} = 90 \,km/h$ for both datasets). It can be seen that, on the roads with different speed limits, both traffic volume and traffic speed can vary significantly, as particular speed limit values can impose a notable restriction on the permitted traffic speed.

\clearpage

The next Figure \ref{fig:targets-dynamics-crosswalk} presents the target dynamics for the road subsets with different values of the \texttt{ends\_with\_crosswalk} feature. When moving on the roads that end with crosswalks, drivers have to slow down their vehicle in order to let pedestrians pass, which significantly affects the average traffic speed registered on such roads and makes it much lower on average.

\begin{figure}[h]
    \centering
    \begin{subfigure}{\linewidth}
    \centering
    \includegraphics[width=\linewidth]{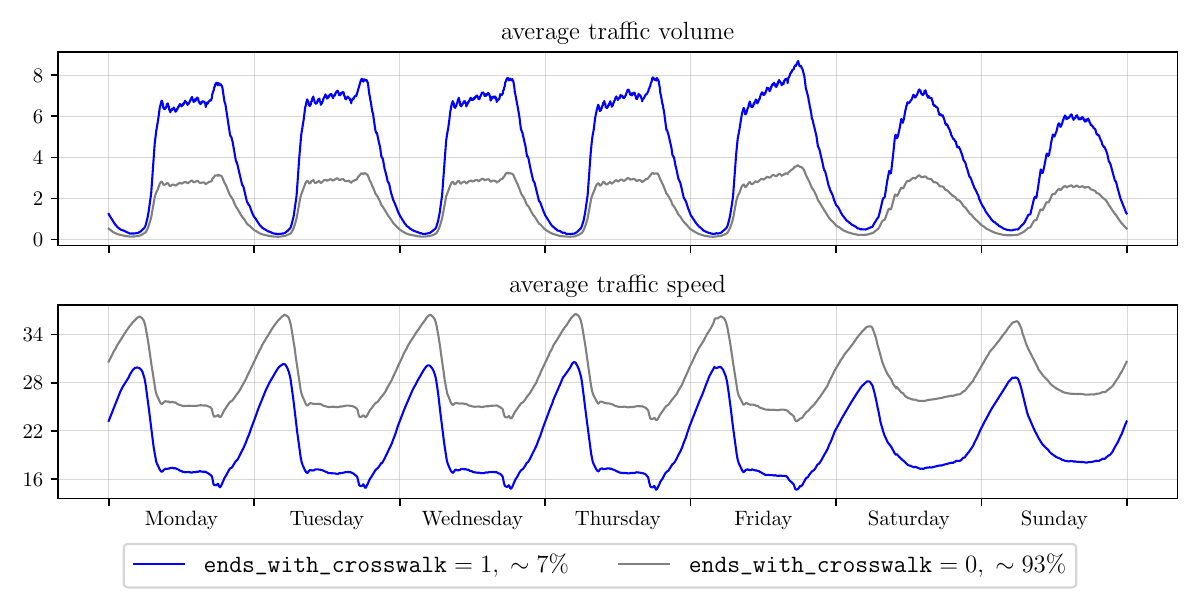}
    \caption{\texttt{city-traffic-M}}
    \end{subfigure}
    \begin{subfigure}{\linewidth}
    \centering
    \includegraphics[width=\linewidth]{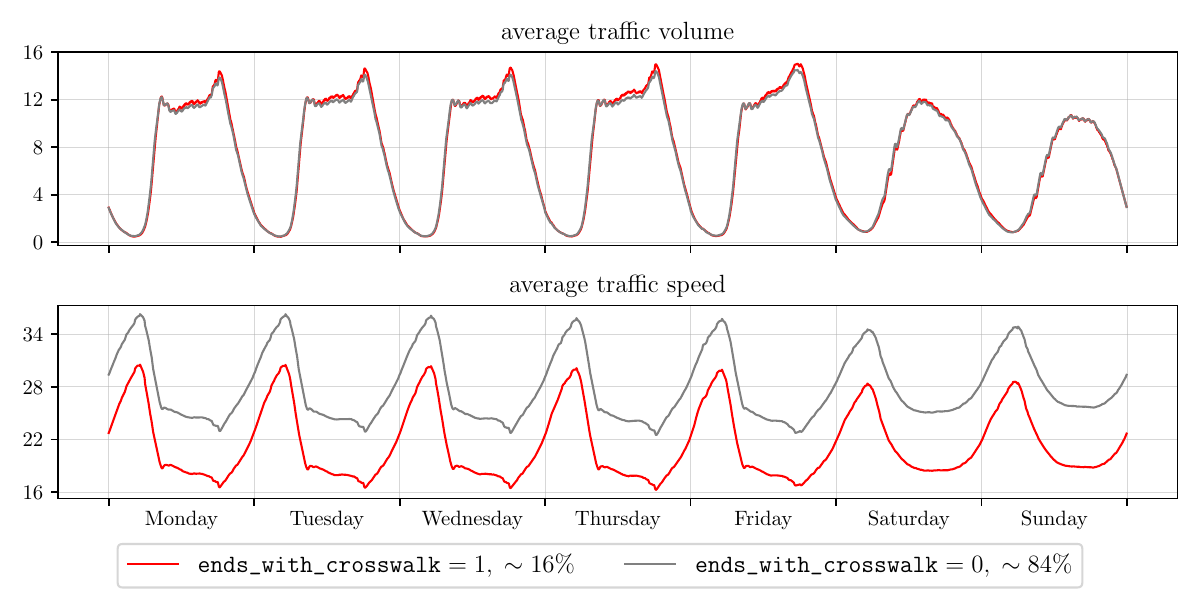}
    \caption{\texttt{city-traffic-L}}
    \end{subfigure}
    \caption{The weekly target dynamics averaged across different road subsets depending on the value of \texttt{ends\_with\_crosswalk}. The percentage in the legend denotes the fraction of nodes in the corresponding category.}
    \label{fig:targets-dynamics-crosswalk}
    \vspace{-20pt}
\end{figure}

\clearpage

\begin{figure}[h]
    \centering
    \begin{subfigure}{\linewidth}
    \centering
    \includegraphics[width=\linewidth]{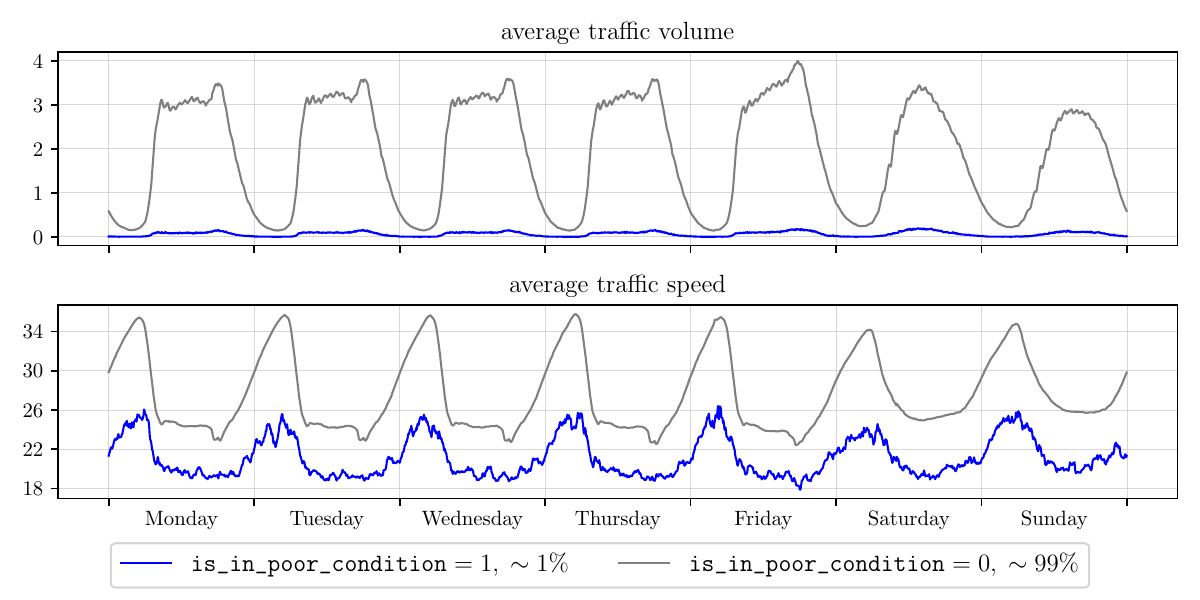}
    \caption{\texttt{city-traffic-M}}
    \end{subfigure}
    \begin{subfigure}{\linewidth}
    \centering
    \includegraphics[width=\linewidth]{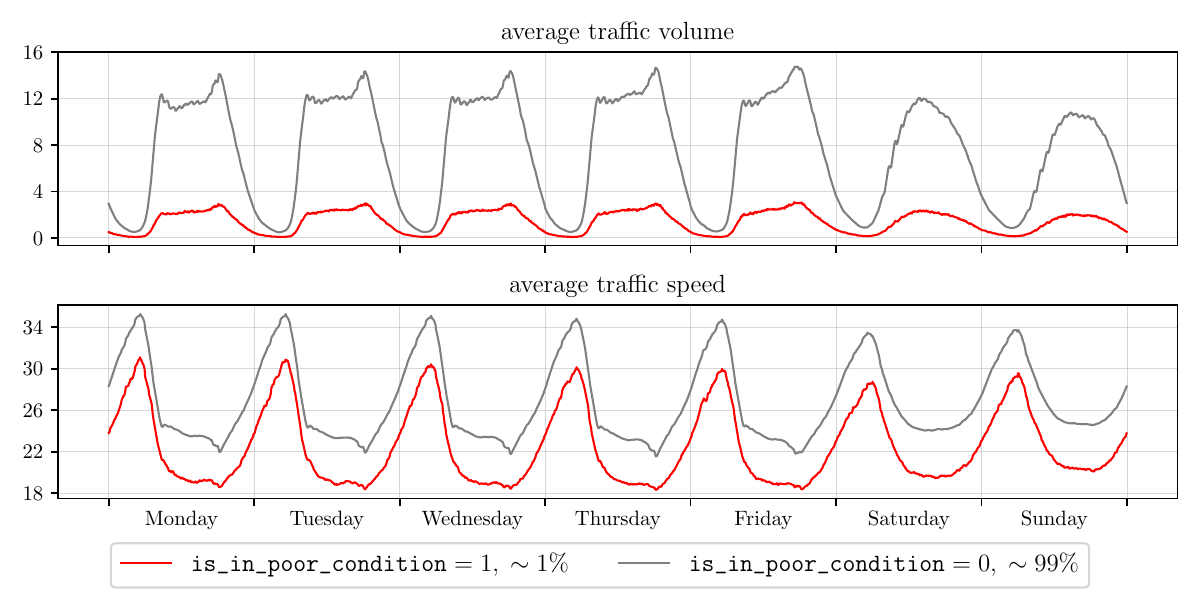}
    \caption{\texttt{city-traffic-L}}
    \end{subfigure}
    \caption{The weekly target dynamics averaged across different road subsets depending on the value of \texttt{is\_in\_poor\_condition}. The percentage in the legend denotes the fraction of nodes in the corresponding category.}
    \label{fig:targets-dynamics-condition}
\end{figure}

In Figure \ref{fig:targets-dynamics-condition}, we show the dynamics of targets variables for the subsets of roads that have different values of the \texttt{is\_in\_poor\_condition} feature. If a road is in poor condition, drivers have to move on it more carefully and keep speed low in order to avoid any accidents. At the same time, there are not so many such roads in both cities, so traffic volume on the roads with normal condition is much higher on average.

\clearpage

\begin{figure}[h]
    \centering
    \begin{subfigure}{\linewidth}
    \centering
    \includegraphics[width=\linewidth]{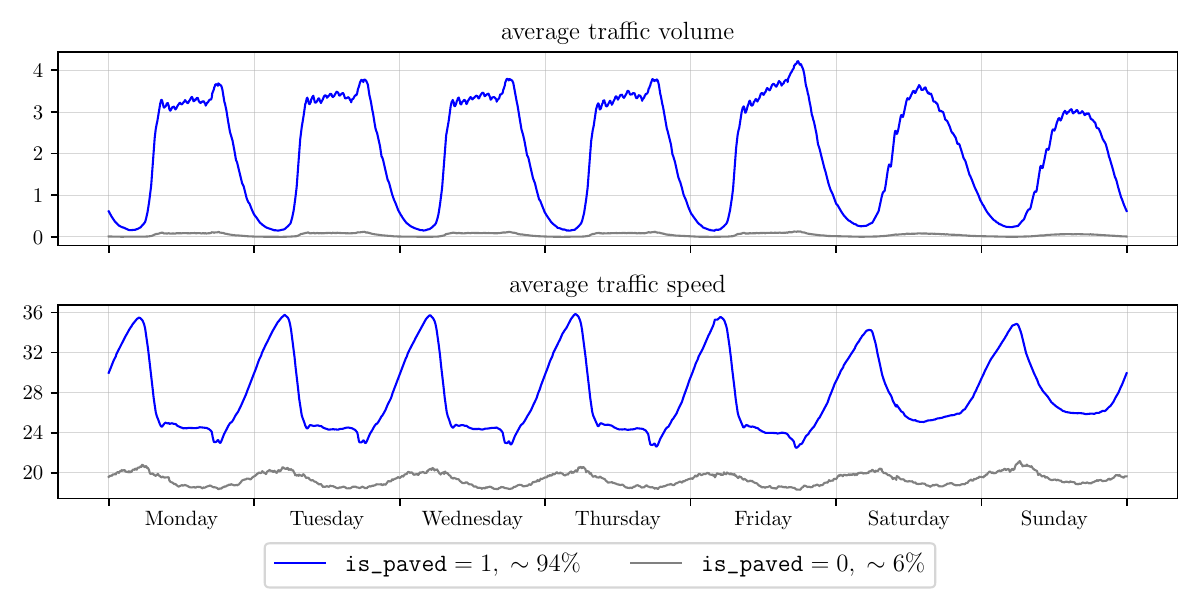}
    \caption{\texttt{city-traffic-M}}
    \end{subfigure}
    \begin{subfigure}{\linewidth}
    \centering
    \includegraphics[width=\linewidth]{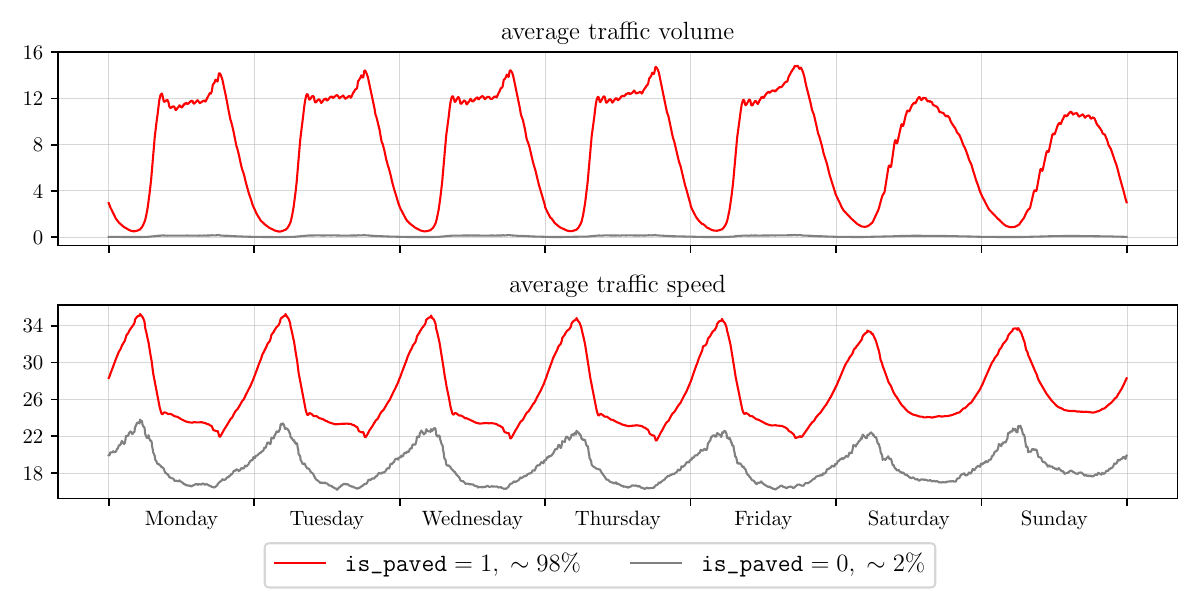}
    \caption{\texttt{city-traffic-L}}
    \end{subfigure}
    \caption{The weekly target dynamics averaged across different road subsets depending on the value of \texttt{is\_paved}. The percentage in the legend denotes the fraction of nodes in the corresponding category.}
    \label{fig:targets-dynamics-paved}
\end{figure}

Figure \ref{fig:targets-dynamics-paved} presents the target dynamics for the roads with different value of the \texttt{is\_paved} feature. The movement on paved roads is more convenient and fast, which leads to higher traffic speed on average. Also, since pavement is a standard in road construction nowadays, the majority of roads in both cities have necessary surface, and most traffic volume is distributed exactly over paved roads.

\clearpage

\begin{figure}[h]
    \centering
    \begin{subfigure}{\linewidth}
    \centering
    \includegraphics[width=\linewidth]{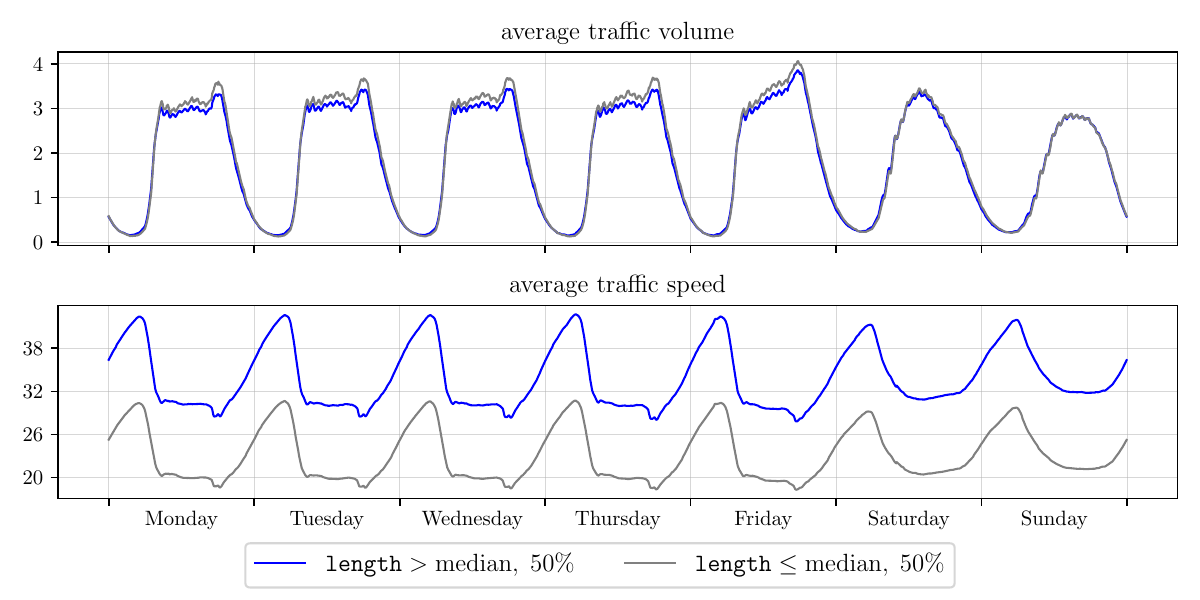}
    \caption{\texttt{city-traffic-M}}
    \end{subfigure}
    \begin{subfigure}{\linewidth}
    \centering
    \includegraphics[width=\linewidth]{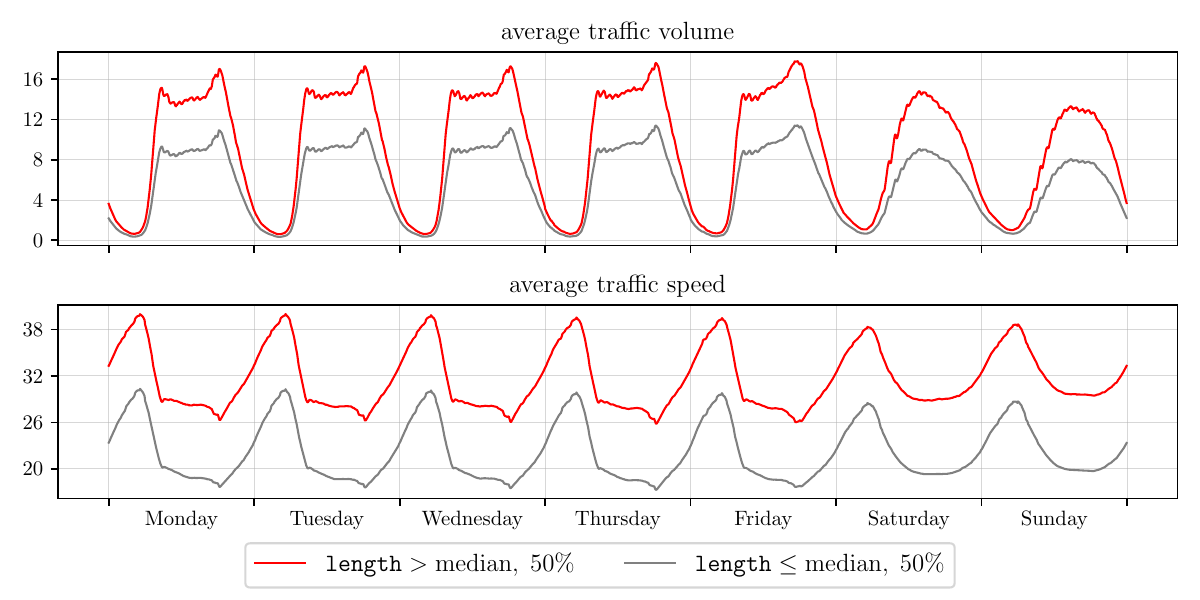}
    \caption{\texttt{city-traffic-L}}
    \end{subfigure}
    \caption{The weekly target dynamics averaged across different road subsets depending on the value of \texttt{length}. The percentage in the legend denotes the fraction of nodes in the corresponding category.}
    \label{fig:targets-dynamics-length}
\end{figure}

In Figure \ref{fig:targets-dynamics-length}, we show the dynamics of targets variables across the roads with different values of the \texttt{length} feature. It is natural that on longer roads, drivers can afford moving on higher speed, in contrast to short roads that can connect different crossroads and crosswalks and may require to constantly slow down the vehicle. Moreover, since longer roads cover greater distance and typically connect locations with different logistic purpose in the larger city of \texttt{city-traffic-L}, they tend to carry more traffic volume.

\clearpage

\begin{figure}[h]
    \centering
    \begin{subfigure}{\linewidth}
    \centering
    \includegraphics[width=\linewidth]{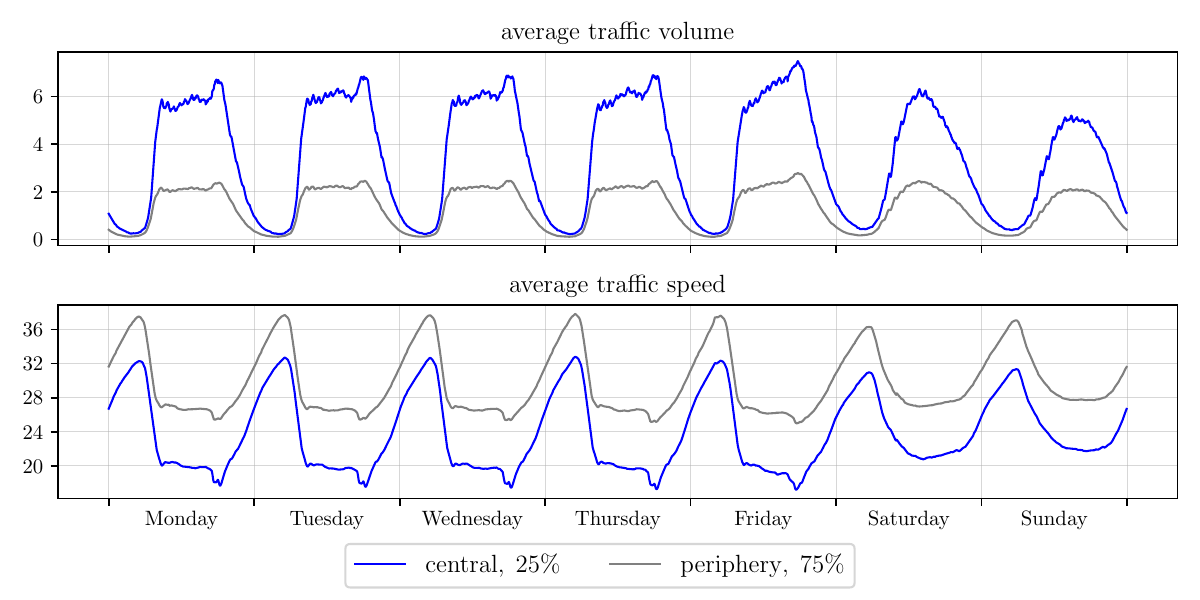}
    \caption{\texttt{city-traffic-M}}
    \end{subfigure}
    \begin{subfigure}{\linewidth}
    \centering
    \includegraphics[width=\linewidth]{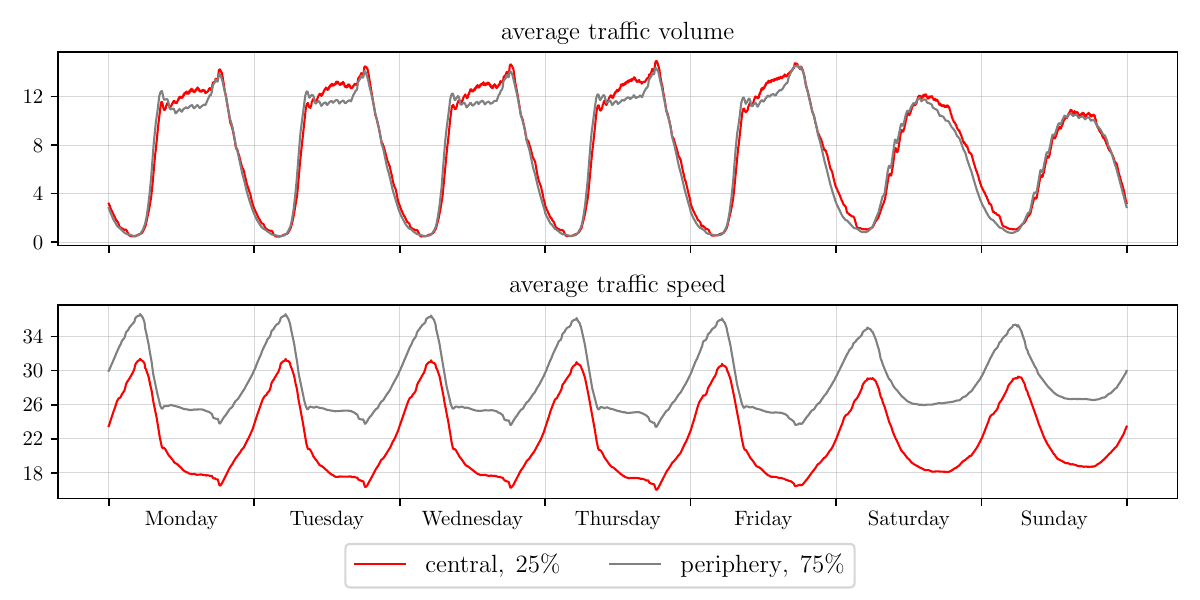}
    \caption{\texttt{city-traffic-L}}
    \end{subfigure}
    \caption{The weekly target dynamics averaged across different road subsets depending on whether they are located at the city center.}
    \label{fig:targets-dynamics-central}
\end{figure}

Figure \ref{fig:targets-dynamics-central} shows the target dynamics for the roads belonging to the central part of city (in our case, we decide to choose $25\%$ of the roads) and to its periphery. In the city center, the structure of road network can be more complex and require more maneuvers to pass through it, so the average traffic speed on the central roads appears lower than on the peripheral ones. Further, since the city center in the smaller city of \texttt{city-traffic-M} has a more developed and diverse infrastructure that serves various needs of city residents, there is naturally more traffic volume.

The presented figures show that our proposed datasets contain important spatial information about road networks that has strong connection with the traffic speed and volume and thus is necessary to be used for precise traffic forecasting.

\clearpage

\section{Experimental setup}
\label{app:experimental-setup}

In all our experiments, we use a lookback window of 48 timestamps and train all models to predict next 12 timestamps of the temporal component.

We use learnable node embeddings for road segments in addition to their static features. We also use additional temporal calendar features such as day of the week and timestamp (hour and minute) during the day. We encode these features both with one-hot encoding and with periodic trigonometric functions.

To ensure a fair comparison between models with substantially different computational costs, we impose a fixed training budget of 12 hours for each model and select the configuration that achieves the best validation MAE within this budget. Since preliminary experiments showed that the learning rate and hidden dimension have the strongest effect on performance, we tune these two hyperparameters. In particular, we search over hidden dimensions from $\{64, 128, 256\}$ and learning rates from $\{0.0003, 0.001, 0.003, 0.01\}$. This setup allows each model to use the available computational budget in the most effective way. The selected configurations are reported in Table~\ref{tab:selected-hparams}.

To ensure comparability across experiments, we fix the effective batch size to $30$ across all datasets and adjust gradient accumulation steps as needed. In Table~\ref{tab:model-performance}, we report the mean and standard deviation (over multiple random seeds) of test Mean Absolute Error (MAE).

All experiments are constrained to a single A100 GPU with 80GB of VRAM and 120GB of system RAM, and conducted in a full-batch training mode without neighbor sampling. This choice is motivated by the need for consistent and fair comparison between models, particularly because neighbor sampling introduces stochasticity that can disproportionately affect certain architectures and complicate evaluation. Moreover, given the scale of our datasets and the memory available on a single GPU, full-batch training remains feasible and provides deterministic gradient computations that improve stability and reproducibility.

We use \texttt{dgl==2.4.0+cu124} and \texttt{torch==2.4.0+cu124} for our experiments.

\begin{table}[h!]
\centering
\caption{Selected hyperparameter configurations based on the best validation MAE within the 12-hour training budget.}
\vspace{2pt}
\label{tab:selected-hparams}
\begin{tabular}{lcccc}
\toprule
& \multicolumn{2}{c}{\texttt{city-traffic-M}} & \multicolumn{2}{c}{\texttt{city-traffic-L}} \\
\cmidrule(lr){2-3} \cmidrule(lr){4-5}
Model & \texttt{lr} & \texttt{hidden dim} & \texttt{lr} & \texttt{hidden dim} \\
\midrule
GRUGCN & $0.003$ & $256$ & $0.003$ & $64$ \\
DCRNN & $0.003$ & $64$ & $0.003$ & $64$ \\
STGCN & $0.003$ & $64$ & $0.003$ & $64$ \\
GWN & $0.003$ & $64$ & $0.003$ & $64$ \\
SVR-GNN & $0.001$ & $256$ & $0.001$ & $256$ \\
\bottomrule
\end{tabular}
\end{table}

\clearpage

\section{Details of the proposed efficient spatiotemporal GNN}
\label{app:architecture}

In this section, we describe in more detail the efficient neural network architecture of SVR-GNN.

Let $x_i \in \mathbb{R}^F$ be the input feature vector for node $i$ with static node features (road segment coordinates, road segment speed limit, etc; see Appendix~\ref{app:datasets-details} for more details), learnable node embeddings, temporal calendar features (day of the week, etc.), and past values of traffic speed and volume in the lookback window (all these features are concatenated in a single vector of dimension $F$). First, our architecture transforms this input feature vector into a single hidden representation $h_i^0 \in \mathbb{R}^H$ (where $H$ is the hidden dimension) with a linear layer, followed by dropout \citep{srivastava2014dropout} and a GELU activation function \citep{hendrycks2016gaussian}:
\[ h_i^0 = \texttt{GELU}(\texttt{Dropout}(\texttt{Linear}(x_i))). \]
Then, this hidden representation is iteratively transformed by $L$ GNN blocks:
\[ h_i^{l+1} = \texttt{GNNBlock}(h_i^l). \]
Finally, the hidden representation obtained from the last GNN block is normalized with a layer normalization \citep{ba2016layer} and transformed to predictions $\widehat{y}_i$ with a linear layer:
\[ \widehat{y}_i = \texttt{Linear}(\texttt{LayerNorm}(h_i^L)). \]
Our GNN block is based on the architecture from \citet{platonov2023critical, GraphLand}, and, besides a neighborhood aggregation operation, includes a residual connection \citep{he2016deep}, a layer normalization, and a 2-layer MLP. We also concatenate the result of neighborhood aggregation with the representation of the aggregating node, similar to GraphSAGE \citep{hamilton2017inductive}. Let $N(i)$ be the set of 1-hop neighbors of node $i$ in the graph. Then, a single GNN block can be described as follows:
\[
h_i^{\text{norm}} = \texttt{LayerNorm}\big(h_i^{\text{input}}\big),
\]
\[
\vspace{2pt}
h_i^{\text{aggr}} = \texttt{Concatenate}\left[h_i^{\text{norm}}, \texttt{Mean}\Big(\{ h_j^{\text{norm}} \; \forall j: j \in N(i) \}\Big)\right],
\]
\[
\vspace{2pt}
h_i^{\text{output}} = h_i^{\text{input}} + \texttt{MLP}\left(h_i^{\text{aggr}}\right).
\]

\section{Code and data accessibility}
\label{app:code-and-data}

Our datasets are shared via \href{https://www.kaggle.com/datasets/mightyneghbor/city-traffic-benchmarks/data}{Kaggle page}, and our code can be accessed in \href{https://github.com/yandex-research/urban-traffic-benchmark}{GitHub repository}.

\end{document}